%% file: AAAAAbarejrnl.tex
\RequirePackage{fix-cm}
\documentclass[twocolumn]{svjour3}          % twocolumn
\smartqed  % flush right qed marks, e.g. at end of proof
\usepackage{epsfig}
\usepackage{graphicx}
\usepackage{amsmath}
\usepackage{amssymb}
\usepackage{subfiles}
\usepackage{multirow}

\usepackage{booktabs}
\input{math_commands}

\usepackage{algorithm}
\usepackage{algorithmic}

\usepackage[utf8]{inputenc}
\usepackage[english]{babel}
\usepackage{xspace}
\makeatletter
\DeclareRobustCommand\onedot{\futurelet\@let@token\@onedot}
\def\@onedot{\ifx\@let@token.\else.\null\fi\xspace}
 
\def\ie{\emph{i.e}\onedot}

\def\etal{\emph{et al}\onedot} 
\makeatother

\usepackage{makecell}

\usepackage{xcolor}
\usepackage{color, colortbl}
\definecolor{citecolor}{HTML}{0071bc}
\definecolor{tabhighlight}{HTML}{e5e5e5}
\usepackage[colorlinks,citecolor=citecolor]{hyperref}

\usepackage{multirow} 
\usepackage{comment}
\usepackage{booktabs}
\usepackage{wrapfig}
\usepackage{subfig}
\usepackage{bbm}
\usepackage{amssymb}
\usepackage{pifont}

\begin{document}
\sloppy

\title{TripleE: Easy Domain Generalization via Episodic Replay}

\author{Xiaomeng Li \and
        Hongyu Ren \and
        Huifeng Yao \and
        Ziwei~Liu}

\institute{Xiaomeng Li \at
              The Hong Kong University of Science and Technology \\
              \email{eexmli@ust.hk}           %  \\
           \and
          Hongyu Ren \at
              Stanford University \\
              \email{hyren@cs.stanford.edu}
           \and
           Huifeng Yao \at
              The Hong Kong University of Science and Technology \\
         \email{hyaoad@connect.ust.hk}
           \and
           Ziwei Liu \at
           S-Lab, Nanyang Technological University, Singapore \\
           \email{ziwei.liu@ntu.edu.sg}
}

\date{Received: date / Accepted: date}

\maketitle

\newcommand{\xmli}[1]{{\color{blue}{[XM: #1]}}}

\input{sections/1.Abstract.tex}
\input{sections/2.Introduction.tex}

\input{sections/3.Related_work.tex}

\input{sections/4.Methodology.tex}
\input{sections/5.Experiments.tex}

\input{sections/6.Conclusion.tex}

{
\bibliographystyle{plain}
\bibliography{sections/bibliography}
}

\end{document}

%% file: math_commands.tex
\usepackage{amsmath,amsfonts,bm}

\def\eqref#1{equation~\ref{#1}}
\def\1{\bm{1}}

\def\vx{{\bm{x}}}

\DeclareMathAlphabet{\mathsfit}{\encodingdefault}{\sfdefault}{m}{sl}
\SetMathAlphabet{\mathsfit}{bold}{\encodingdefault}{\sfdefault}{bx}{n}

\def\gA{{\mathcal{A}}}
\def\gB{{\mathcal{B}}}

\def\gD{{\mathcal{D}}}

\def\gF{{\mathcal{F}}}

\def\gP{{\mathcal{P}}}

\def\gT{{\mathcal{T}}}

%% file: sections/1.Abstract.tex
\begin{abstract}
% This paper presents a frustratingly easy yet novel method given the recent progress in domain generalization.
Learning how to generalize the model to unseen domains is an important area of research.
%We discover a key insight for domain generalization is that a good model should learn robust and consensual features and not be affected by biased or noisy samples in the training source domains.
In this paper, we propose TripleE, and the main idea is to encourage the network to focus on training on subsets (learning with replay) and enlarge the data space in learning on subsets.
Learning with replay contains two core designs, EReplayB and EReplayD, which conduct the replay schema on batch and dataset, respectively. Through this, the network can focus on learning with subsets instead of visiting the global set at a glance, enlarging the model diversity in ensembling. 
To enlarge the data space in learning on subsets, we verify that an exhaustive and singular augmentation (ESAug) performs surprisingly well on expanding the data space in subsets during replays.
Our model dubbed \textbf{TripleE} is frustratingly easy, based on simple augmentation and ensembling.
Without bells and whistles, our TripleE method surpasses prior arts on six domain generalization benchmarks, showing that this approach could serve as a stepping stone for future research in domain generalization.
\keywords{Domain generalization \and Augmentation \and Learning with replay }
\end{abstract}

%% file: sections/2.Introduction.tex
\section{Introduction}
%\hongyu{to be honest, I feel you used frustratingly too frequently. It won't be appreciated by the reviewers}
%\hongyu{my comments on the abstract also seem to apply in the intro}
% background 
Deep neural networks have achieved remarkable success on various computer vision tasks~\cite{garcia2017review,li2017fully,huang2017densely,zhou2018unet++,li2018h} with an assumption that the training and test datasets consist of \textit{i.i.d.} samples from the same distribution. However, in real-world scenarios, the test data (target domain) are often outside the training dataset domains (source domains), posing a significant challenge to deep learning algorithms. Domain generalization (DG) assumes that a model is trained from multiple source domains and expected to perform well on unseen domains. In such scenarios, the model inherently cannot capture any distribution shift between the source and target domains. Thus, a robust training method is crucial to domain generalization tasks.

% Considerable efforts have been devoted to developing DG methods. One line of research proposed DG methods with the consideration of domain labels.

%To address the limitation, various domain adaptation methods ~\cite{ganin2016domain,hoffman2018cycada,kumar2010co,saito2018maximum,tzeng2015simultaneous,tzeng2017adversarial} minimized the domain discrepancy by aligning the source and target domain distributions.  
%These domain adaptation methods require access to the data from target domains during training. However, in real-world applications, it is often infeasible to obtain such target domain data at the training stage~\cite{li2018domain,yue2019domain}.
%\hr{do they try to generate a new source domain or generate new data within the given source domains?}. 
%For instance, Volpi~\etal~\cite{volpi2018generalizing} developed an approach based on adversarial learning~\cite{goodfellow2014generative}, where additional datapoints are generated by perturbation according to fictitious target distributions within a certain Wasserstein distance from the source;

Recent studies on DG have explored several directions, and these methods can be broadly classified into three categories: 
1) Learning domain-invariant representation~\cite{wang2020learning,dou2019domain,Carlucci_2019_CVPR,li2018domain}, which aims to learn a shared domain-invariant representation for images from multiple source domains.
2) General model regularization~\cite{huang2020self,shi2020informative,wang2019learning}, which imposes additional regularization terms upon the model during training, \emph{e.g.}, gradient dropouts~\cite{huang2020self}, informative dropout~\cite{shi2020informative} and superficial information regularization~\cite{wang2019learning}.
3) Augmenting source domains~\cite{volpi2018generalizing,somavarapu2020frustratingly,zhou2020learning,li2021simple,xu2021fourier}, whose goal is to enlarge the training data to a broader span of the data space, increasing the possibility of covering the data in the target domain.
%These methods can also be classified into two categories according to whether they use domain labels or not; see the partition in Table~\ref{tab:PACS_res18}. 
Although prior works have achieved promising results, most methods are complicated, containing hand-crafted architectures, memory banks, multiple training stages, or self-supervised tasks~\cite{Carlucci_2019_CVPR,wang2020learning,chen2021style}. %For example, representation-based methods~\cite{Carlucci_2019_CVPR,wang2020learning} designed self-supervised tasks and combined multiple loss functions to learn robust features for domain generalization. 
For example, Chen~\etal~\cite{chen2021style} proposed a style and semantic memory mechanism, which requires building multiple memory banks to store style and semantic features. 
Xu~\etal~\cite{xu2021fourier} proposed a Fourier transformation for DG based on a teacher-student model.

Unlike prior work, we propose a frustratingly easy and powerful baseline method for domain generalization with simple augmentation and ensembling. 
% We observe that large batch training often results in degraded generalization, which is caused by a tendency of low gradient variance updates to converge to ``sharp minima''~\cite{keskar2016large}. Some pieces of evidence showed that a decrease in gradient variance reduction in each optimization step could help the model avoid sharp minima and reach a more robust solution~\cite{hoffer2020augment}. 
We discover a key insight of DG is to encourage the network to \textbf{\textit{focus on training on subsets (learning with replay)}} and \textbf{\textit{enlarge the data space in performing replays}}.  
% The main idea is to encourage the network \textbf{\emph{to focus on training on subsets (learning with replay) and enlarge the data space in performing replays.}} 
The \emph{subsets} should belong to the whole set and can be defined on \emph{batch} and on \emph{dataset} level. Therefore, learning with replay has two core designs: \textbf{1)} Episodic replay on batch (\textbf{EReplayB}): augment each batch multiple times. By augmenting \emph{batch} multiple times, EReplayB encourages the network to focus on learning with \emph{subsets (batch).}
%We treat each batch as a \emph{subset} and EReplayB can encourage the network train \emph{subsets} multiple times. 
\textbf{2)} Episodic replay on dataset (\textbf{EReplayD}): randomly split the dataset into sub-datasets, train models on sub-datasets individually, periodically resplit sub-datasets, and ensemble the predictions of models trained with sub-datasets. 
To enlarge the data space in learning on \emph{subsets}, we further introduce \textbf{3)} Exhaustive, and singular augmentation (\textbf{ESAug}): a cascade of successive compositions can produce images that drift far from the original image and lead to unrealistic images. Unlike prior work, we employ exhaustive and singular augmentation to enhance the generalization.

% a singular augmentation randomly selected from an exhaustive augmentation list performs surprisingly well on expanding the data space in performing replays. Our completed model called \textbf{TripleE} do not need any additional module design. 

Overall, EReplayB and ESAug work well because they can effectively decrease gradient variance reduction in each optimization step, could help the model avoid sharp minima, and reach a more robust solution. 
EReplayD enforces the model to learn with sub-datasets instead of visiting the whole set at a glance, and the model has the chance to visit the whole set because of periodic updates of sub-datasets. Therefore, it can generate more diverse models and is demonstrated to be more effective than the traditional ensembling method. 
% ~\cite{hoffer2020augment}

%Training on subsets can increase the gradient variances, avoiding the model converge to sharp minima; While enlarging the data space in performing replays can generate more correlated gradients within the batch, leading to a reduced gradient variance. 

% Specifically, EReplayB augments
% each image multiple times. Instead of learning the whole dataset directly, EReplayD trains a single model on \textbf{\textit{periodically-updated subsets}} and eventually is exposed to
% the entire dataset. ESAug achieves \textbf{\textit{enlarge the data space in performing replays}} by an exhaustive and singular augmentation. 

%Our idea shares a similar spirit with RANSAC~\cite{fischler1981random}, which estimates a mathematical model from dataset containing outliers by randomly sampling observed data and can be interpreted as an outlier detection method. 
% Similarly, an ideal DG model should learn robust and consensual features and should not be affected by \emph{biased or noisy} samples in the training source domains. 
% To this end, we propose an augmentation and ensembling approach, and the

%Our model dubbed \textbf{TripleE} is frustratingly simple and flexible, based on simple augmentations and ensembling without any particular loss functions, model, and structure design.  

The main contributions of this paper are listed as follows: 
\begin{itemize}
%\item We present a frustratingly easy baseline method (TripleE) that largely outperforms prior art and could serve as a stepping stone for future research in domain generalization.

\item We reveal a key secret for the performance gains is to encourage the network to train with subsets and enlarge the data space in learning on subsets. This simple but overlooked point should be known in the fast-growing DG area. 

\item We propose TripleE, which consists of three components: EReplayB, EReplayD, and ESAug. EReplayB and EReplayD perform \emph{learning with replay} on batch and dataset, respectively. ESAug can effectively \emph{enlarge the data space during replays}. 

%We reveal a key secret for the performance gains is to encourage the network to train subsets and enlarge the data space in learning subsets. Therefore, we derive TripleE based on simple augmentation and ensembling. 
\item Without bells and whistles, TripleE outperforms other state-of-the-art methods on six DG benchmarks, notably achieving 5.54\% and 2.23\% improvements on Digits-DG and PACS, respectively, showing that our method could serve as a stepping stone for future research in domain generalization~\footnote{Code is available at \url{https://github.com/xmed-lab/TripleE-DG}}. 
% We demonstrate a randomized model ensemble method can greatly enhance the model's generalizability. 
% \item Without bells and whistles, our approach surpasses all previous state-of-the-art results on Digits-DG, PACS, and OfficeHome datasets. We hope this frustratingly easy baseline method 
\end{itemize}

\if 
Learning category-aware domain-invariant representation from source domains is one crucial key to success in DG; 
The insight here is to learn generalizable features across multiple source domains by pulling together all the images with the same labels (\emph{positives}) and pushing away images with different labels (\emph{negatives}) in the embedding space. 
The insight shares a similar spirit with contrastive learning, which was developed for unsupervised representation learning~\cite{chen2020simple,he2020momentum}.  
However, directly employing contrastive learning to DG yields limited performance, and there are several reasons. 
Contrastive learning requires a large batch size~\cite{chen2020simple} or memory bank~\cite{he2020momentum} to construct sufficient \emph{positives} and \emph{negatives} to learn representation from a large amount of data. 
Instead of learning representations from a large amount of data, DG aims to improve the model generalization on unseen target domains with a distribution gap from the source domains used in training.
Previous studies have demonstrated that large-batch training can hurt the model generalization especially when there exists a distribution shift between training and test~\cite{masters2018revisiting,keskar2016large,hoffer2020augment}. This is exactly the assumption in DG problems and we also observe this phenomenon in the experiment; see Figure~\ref{fig:f2}.  
% Moreover, DG datasets used in the existing literature have very limited amounts. For example, PACS dataset~\cite{li2017deeper} contains 9,991 images and is largely smaller than the dataset used for unsupervised representation learning. 
Besides, DG problems contain training images from multiple source domains and have significant image variations (even among positives) than unsupervised learning, which requires careful reweighting and modeling. Hence, increasing the generalization of contrastive learning using small-batch training among multiple source domains provides a promising direction for advancing domain generalization performance.

%% nour novelty 
To this end, we present a new paradigm called Instance-repeated Domain Contrast (IDC) that learns domain-agnostic category-aware representation via a domain contrast loss with instance-repeated small-batch training.
First, we formulate a domain contrast loss to optimize \emph{positives and negatives} across domains in a unified formulation, with the goal of encouraging the encoder to push together positives and separate negatives in the embedding space. 
Second, we adapt domain contrast to small-batch training to take the generalization benefits of small-batch.
However, directly integrating two techniques is problematic since domain contrast requires sufficient positive and negative samples to learn representation, whereas %small-batch training makes the existence of multiple samples with the same label in one batch less likely. 
small-batch training limits the sample numbers during each feedforward calculation. 
Hence, we further introduce instance repetition to randomly augment each sample multiple times to create adequate samples within a batch but keeping the statistics of image semantics unchanged. 
By instance repetition, we still maintain a small core batch of images. 
We then integrate domain contrast with small-batch instance repetition in a unified framework, which provides highly robust feature representation as well as takes the generalization benefits of small-batch training.
We demonstrate the effectiveness of our approach on three benchmark datasets with ablation studies. 

The main contributions of this work are summarized as follows:
\begin{itemize}
	
	\item %We first propose domain contrast, a unified formulation to optimize \emph{all positives} and \emph{negatives} in the embedding space, where \emph{positives} include images within and across domains. 
	We propose a contrastive learning solution to solve the domain generalization problem and it is is a novel research direction for DG.
%	Our method is simple, straightforward and effective. 
Our method is simple, and effective and can serve as a plug-and-play module to be combined with and improve other DG methods. 
	
	\item  We design a novel contrastive learning with instance-repeated small-batch training for the first time. Our method simultaneously learns the domain-agnostic representation and leverages the generalization benefits of small-batch training.

	\item Our IDC achieved the best performance on two commonly used benchmark datasets, significantly outperforming the other state-of-the-arts by 3.6\% and 4.8\% relative improvements on the PACS and Digits-DG datasets, respectively. 
	
	\item The ability to generalize across domains is a key challenge not only in computer vision, but also in medical imaging field. We introduced the first medical image dataset for domain generalization,~\ie, COVID-19 X-ray dataset. 
	Our code and data will be publicly available on GitHub upon acceptance.

\end{itemize}
\fi 

\if 1 
Different from prior work, we provide a novel direction for DG based on the following key insights.
First, previous studies have demonstrated that small-batch training can significantly increase network robustness, especially when training data are scarce~\cite{masters2018revisiting,keskar2016large,hoffer2020augment}. 
This aligns well with the domain generalization setting, where we do not have much training data from each source domain and thus effectively avoiding overfitting is essential to generalization performance on the unseen domain. 
However, this factor is ignored in the existing methods and requires careful design to be applicable in a multi-source domain setting. 
Second, learning category-aware domain-invariant representation from source domains is one crucial key to success in DG; 
The insight here is to pull together all the images with the same labels across domains (\emph{positives}), and push away images with different labels (\emph{negatives}) in the embedding space.  
However, how to encourage the network to learn such representation by small-batch training and by considering the distances in the embedding space among \emph{all positives and negatives} without using domain partition labels remains unknown. 
To this end, we present a new paradigm called Instance-repeated Domain Contrast (IDC) that learns domain-agnostic category-aware representation via a domain contrast loss with instance-repeated small-batch training.
First, we formulate a domain contrast loss to optimize \emph{all the positives and negatives} across domains in a unified form, with the goal to encourage the encoder to push together positives and separate negatives in the embedding space. 
Second, we adapt domain contrast to small-batch training in order to take the generalization benefits of small-batch.
However, directly integrating two techniques is problematic since domain contrast requires multiple positive samples to learn features, whereas small-batch training makes the existence of multiple samples with the same label in one batch less likely. 
Hence, we further introduce instance repetition to randomly augment each sample multiple times to create adequate positives within a batch while keeping the statistics of image semantics unchanged. 
By instance repetition, we still maintain a small core batch of images. 
We then integrate domain contrast with small-batch instance repetition in a unified framework, which provides high robust feature representation as well as takes the generalization benefits of small-batch training.

\fi 

%\hr{Reorganize, e.g., start with second and third, and then back to the first (small batch direction). The reason is that you want to introduce your method here, there is no need to start with small-batch training has good result. Achieving good result should come after the method description, you can just empasize again that small batch training helps robustness and avoids overfitting etc.}

%% file: sections/3.Related_work.tex
\section{Related Work}\label{sec:related}

Considerable efforts have been devoted to DG to design models  which reduce reliance on domain-specific artifacts.
In the following, we review the related DG literature from three perspectives. Furthermore, we discuss some related approaches on batch training techniques and augmentations.

\noindent {\bf DG via Learning Domain-invariant Representation.}
The research in this direction aims to learn domain-invariant representation by minimizing the discrepancy between source domains~\cite{ghifary2015domain,motiian2017unified,li2017deeper,li2017learning,Li_2019_ICCV,dou2019domain,li2018domain}. The resulting domain-invariant representation can then be used for downstream tasks in the unseen target domain. Along this track, Ghifary~\cite{ghifary2015domain} designed a multi-task autoencoder, which transfers an image to its related domains and thereby learns robust representation across various domains. 
Another related work is the maximum mean discrepancy-based adversarial autoencoder (MMD-AAE)~\cite{li2018domain}, which applies MMD distance to align the distributions among different domains via adversarial training~\cite{goodfellow2014generative}.
%Methods inspired by meta-learning have also been proposed to tackle the domain generalization problem~\cite{dou2019domain, Li_2019_ICCV}.   
%For example, Li~\etal~\cite{Li_2019_ICCV} developed an episodic training method, which decomposes a network to feature encoder and classifier. 
%Each component interacts with each other, mincing the train-test domain shift during training and improving generalization. 
%Inspired by the Siamese architecture~\cite{chopra2005learning}, 
%Motiian~\etal~\cite{motiian2017unified} presented a multi-task paradigm (CCSA) that integrates a classification and a semantic alignment loss, aiming at predicting high classification accuracy and learning semantically aligned visual domains. 
However, these methods require the source domain partitions. 
%Moreover, CCSA relies on a Siamese network to learn embeddings. It contains
%an anchor with one positive and one negative, and a margin
%is needed to control the degree of separation and alignment
%loss, making it difficult to optimize. However, our method
%formulates many pairs of positives and negatives without a
%margin, enforcing the separation and alignments for all sampled
%pairs, leading to better performance.
% % (3) CCSA optimizes
% both classification loss and feature learning loss. Differently,
% IDC solely relies on the proposed domain contrast
% without classification loss, which is simple and even more
% effective to learn domain-invariant representation.

Recent DG methods forgo the requirement of source domain partitions and directly learn with the mixed domains of training data.  
Some methods designed additional regularizations for learning invariant representation. For example, JiGen~\cite{Carlucci_2019_CVPR} developed a self-supervised task, solving jigsaw puzzles, to capture the invariant information within images from multiple source domains. Instead of only capturing invariance within a single image, Wang~\etal~\cite{wang2020learning} proposed to learn an extrinsic relationship among images across domains to improve generalization. 

% Wang~\etal~\cite{wang2019learning2} further introduced patch-wise adversarial regularization to discard predictive signals such as color and textures and rely on global structures.

\noindent {\bf DG via General Model Regularization.} 
The research in this line considers improving domain generalization through reducing the factor of superficial information on the model prediction~\cite{wang2019learning,wang2019learning2,huang2020self,shi2020informative}. For example, Wang~\etal~\cite{wang2019learning} proposed to reduce the model's reliance on known superficial information such as irrelevant textures and encourage the model to rely more on informative parts. 
Some Dropout-like algorithms~\cite{srivastava2014dropout} are also introduced to solve the DG problem~\cite{shi2020informative,huang2017densely}. Shi~\etal~\cite{shi2020informative} designed an InfoDrop to improve the model generalization by reducing model bias to texture. %Huang~\etal~\cite{huang2020self} proposed to locate and mute the most predictive parts of features maps by gradients, train the network to predict the informative parts, and thus improve robustness. 

\noindent {\bf DG via Augmenting Source Domain.} 
Since no information from the target domain is available during training, some researchers proposed to generate synthetic images derived from multiple source domains to increase the diversity of the training data~\cite{zakharov2019deceptionnet,yue2019domain,shankar2018generalizing,volpi2018generalizing,zhou2020learning,somavarapu2020frustratingly,xu2021fourier}. 
By augmenting the source domain data, the model has a broader range of visible domains and a higher possibility of covering the span of target data. 
For example, Yue~\etal~\cite{yue2019domain} proposed to augment the training data by randomizing the synthetic images with the styles of real images from auxiliary datasets. 
% Xu~\etal~\cite{xu2021fourier} proposed a Fourier-based transform to exchange high-level semnatics 

%Shankar~\etal~\cite{shankar2018generalizing} introduced an additional domain classifier to augment the data based on the classification signal. 
%A recent work \cite{somavarapu2020frustratingly} proposed to randomly stylize the training images through an image stylization network, aiming at reducing network bias to textures/shape and thus improving model generalization. 

Our method belongs to the augmentation-based method. 
Unlike the above methods, we present a frustratingly easy domain generalization based on simple augmentation and ensembling. We point out an important insight for DG via augmentations, that is, learning with replay on \emph{subsets} and enlarging the data space when performing replays. However, this simple yet effective point has been overlooked in prior work and should be known given the recent progress in domain generalization.

\if 1 
\noindent {\bf Metric Learning and Contrastive Learning.}
Our key insight is to learn embeddings such that samples from the same class (even across domains) should be embedded closer and samples from different classes are pushed away. 
The idea is related to methods for metric learning and contrastive learning. 
Triplet loss~\cite{weinberger2009distance} and N-pair loss~\cite{sohn2016improved} are two popular loss functions, which are used to minimize the intra-class variation and increase the inter-class variation. 
The former uses only one positive
and one negative sample per anchor, and the latter uses one positive and many negatives. 
Our method is closely related to the contrastive learning literature~\cite{chen2020improved,he2020momentum,chen2020simple,grill2020bootstrap,oord2018representation}. A contrastive loss has been proposed in~\cite{khosla2020supervised} for standard supervised learning tasks.

The key difference is that these methods are designed for general unsupervised learning while our method focuses on improving model generalization across domains. 
We also summarize the key differences in Table~\ref{tab:addlabel}. 
\fi 

\noindent{\bf Batch Training and Augmentation.}
Outside the literature of domain generalization, many recent works studied several general techniques for improved batch training and augmentation, which go beyond augmenting source domain data~\cite{hoffer2020augment,keskar2016large,masters2018revisiting}. 
For example, Hoffer~\etal~\cite{hoffer2020augment} proposed adding repetitive data to a batch to improve robustness in a supervised learning setting. Many other researchers developed various automated data augmentation methods~\cite{cubuk2019autoaugment,lim2019fast,cubuk2020randaugment,muller2021trivialaugment} to define different augmentation search policies and strategies. 
In contrast, this paper reveals a key idea to generalization improvements but neglected in prior work: learning with replay and enlarging the data space when performing replays. 
% We explain this idea works well since it shares a similar spirit with~\cite{hoffer2020augment} and can improve the generalization by simple augmentation and ensembling. 

%% file: sections/4.Methodology.tex
\section{Our Approach} 
\begin{figure*}[t]
	\centering
	\includegraphics[width=1.0\textwidth]{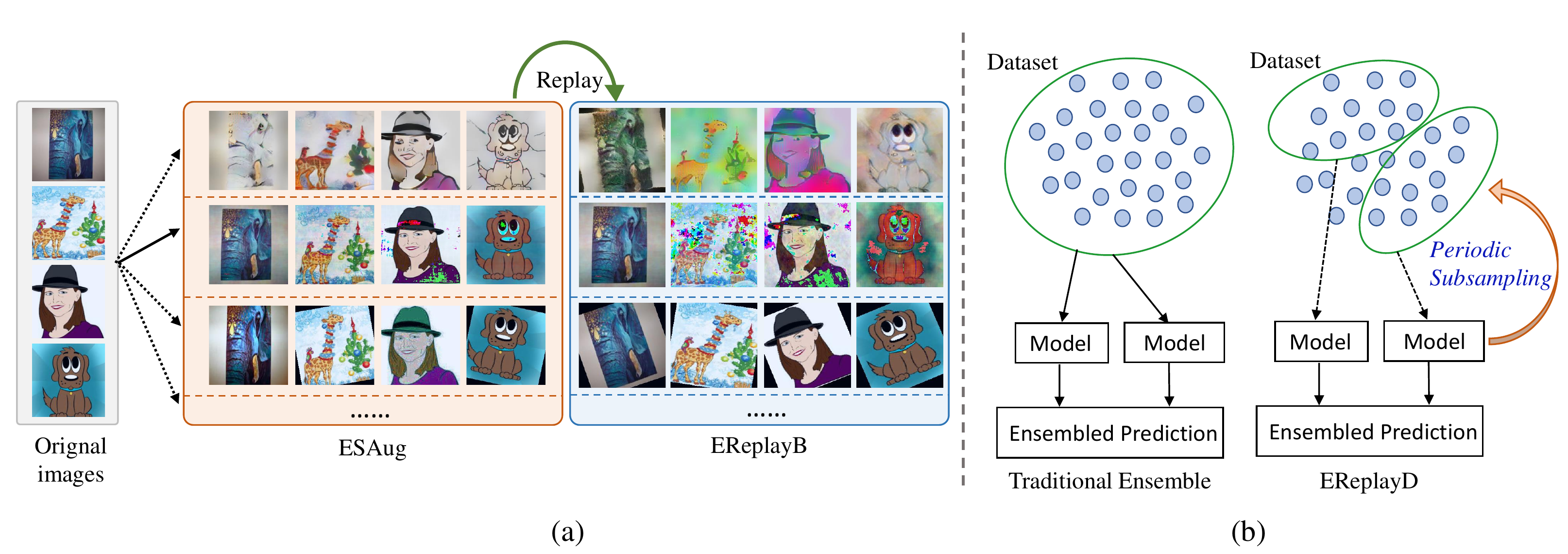}
	\caption{Illustration of TripleE: (a) EReplayB and ESAug. (b) Traditional ensemble vs. EReplayD.} \label{fig:data}
\end{figure*}

We denote $ \mathcal{D} = \left \{ (\vx_i, y_i) \right \}_{i=1}^{N} $ as a mixed training dataset from multiple source domains, consisting of $N$ image-label pairs, where $y_i$ denotes the class label of image $\vx_i$. Note we do not assume we have access to the domain label, \ie, which source domain a given $\vx_i$ belongs to.
The goal in DG is to learn a domain-generalized network $f_{\theta}(\cdot)$ from the source domains and test it on an unseen target domain. Fig.~\ref{fig:data} shows an illustration of TripleE. 
\subsection{EReplayB: Episodically Replay on Batch}
The main idea of EReplayB is to episodically replay each training sample in a batch, and it works as follows. Let $\mathcal{B} \subseteq  \mathcal{D}$ denote a set of  randomly sampled images during training, where the cardinal number of set $\mathcal{B}$ is batch size $b$. We denote  a set of augmentations as $\mathcal{A}$, where $\mathcal{A} = \{a_1, a_2, \dots, a_n\}$. 
The standard operation in many DG methods~\cite{Carlucci_2019_CVPR,xu2021fourier,wang2020learning} is to sequentially apply each augmentation function with some randomness to an input image and the perturbed image can be described as $(a_n...a_3(a_2(a_1(\vx_i))))$.  
We introduce EReplayB to episodically replay each sample in a batch. Specifically, each sample in $\mathcal{B}$ is augmented by $r$ different transformations from $\mathcal{A}$. 
As a result, we can obtain an augmented sample set, denoted as $\mathcal{B}^{'}$ and $|\mathcal{B}^{'}| = r |\mathcal{B}|$. 

As each batch contains multi-views for a single image, it is natural to include positive pairs, which motivates us to leverage the supervised contrastive loss~\cite{khosla2020supervised}. Unlike existing methods designed complicated self-learning loss~\cite{Carlucci_2019_CVPR,wang2020learning}, we use a simple weighted combination of a standard cross-entropy loss and a supervised contrastive loss. We denote $\gP(i)=\{j\in[0,|\gB'|)\:|\:y_j=y_i,\: j\neq i\}$ as the set of indices of all positives (with the same label) in the batch $\gB'$ distinct from $i$ and $\left | \gP(i) \right |$ is its cardinality. Note that we guarantee that $\forall i, \: |\gP(i)| \geq r-1$ since each sample is replayed in the updated batch $\gB'$. 
The positive pairs in the batch include the augmented views of image $\vx_i$ and may also contain other image samples with the same label with image $\vx_i$ from the current or different source domains.
The negative pairs in the batch are images with different labels. 
\begin{equation}
\ell_{sup}  = - \sum_{i}\frac{1}{ \left | \gP(i)\right |}  \sum_{j \in \gP(i)}{\rm log} (\frac{ {\rm exp} (s_{i,j} / \tau ) }{\sum_{k=1}^{r \cdot |\gB|}  \mathbbm{1}_{[k \neq i]} {\rm exp}(s_{i,k}/\tau) }  ), 
\label{eq:1}
\end{equation}
where $\tau>0$ is the temperature; $s_{i,j}, s_{i,k}$ denote the feature similarities among positive pairs and negative pairs, respectively; $r \cdot |\gB|$ is the size of the sample set $\mathcal{B}^{'}$. The total objective is
\begin{equation}
\ell  = \ell_{ce} + \ell_{sup},
\label{eq:2}
\end{equation}
where $\ell_{ce}$ refers to a standard cross-entropy loss. It is worth mentioning that the performance gains of EReplayB are mainly from episodically replaying of the image sample (+4.78\% Acc.), instead of the contrastive loss (+0.74\% Acc.); see results in Table~\ref{tab:ablation_three}. 

%Compared with baseline in other methods~\cite{wang2020learning,xu2021fourier}, our baseline additionally use contrastive loss as a loss function, reaching a higher baseline performance; see comparisons of baselines in Table~\ref{tab:ablation}.  

\subsection{ESAug: Exhaustive and Singular Augmentation}
ESAug enlarges the data space when performing EReplayB and EReplayD. It highlights two core questions: (1) what should be included in the transformation list when performing replays and (2) how to perform augmentation? 

First, we empirically demonstrated that the augmentation should consist of exhaustive transformation functions when learning with replays. We let the augmentation set $\gA$ to include transformations defined in~\cite{cubuk2020randaugment}, each of which can generate a set of distortions to an image. However, these augmentations only perform transformations within a single image. To improve model generalization, we enlarge $\gA$ to $\gA'$ by introducing two cross-image augmentations (Fourier-based augmentation and Style-based augmentation) to encourage the network to learn high-level semantics by exchanging the low-level statistics among images in the source domains.

\noindent \textbf{Fourier-based augmentation.}
Considering that the Fourier phase component contains some semantic-preserving (high-level) information, augmenting the training data by distorting the amplitude information while keeping the phase information unchanged is a good way for the model to learn robust features. 
Through a Fourier-based augmentation, our model can avoid overfitting to low-level statistics carried in the amplitude information, thus paying more attention to the phase information when making decisions.
Specifically, we employ a Fourier-based augmentation by linearly interpolating between the amplitude spectrums of two images from arbitrary source domains:
\begin{equation}
\hat{\gT}(x_i) = (1 - \lambda) \gT(x_i) + \lambda \gT(x_j),   
\end{equation}
where $\gT(x_i)$ refers to amplitude component of image $x_i$ calculated by FFT~\cite{nussbaumer1981fast} algorithm. $\lambda ~\sim  U(0, \xi)$ and hyperparameter $\xi$ controls the strength of the augmentation. The mixed amplitude spectrum is then combined with the original phase spectrum to
form a new Fourier representation:
\begin{equation}
\gF(\hat{x}_i) (u, v) = {\hat \gT}(x_i) (u, v) * e^{{-i} * \gP(x_j)(u, v)},   
\end{equation}
where $\gP(x_i)$ refers to phase component. Then, the augmented  image can be described as 
\begin{equation}
\hat{x}_i = \gF^{-1}[\gF(\hat{x}_i)(u,v)].  
\end{equation}
We denote the set $\gA^f$ to represent the transformation from $x_i$ to $\hat{x}_i$. The final augmentation set can be defined as $\gA'$ and $\gA' = \gA \cup \gA^f$. 
\begin{figure}[!t]
	\centering
	\includegraphics[width=0.48\textwidth]{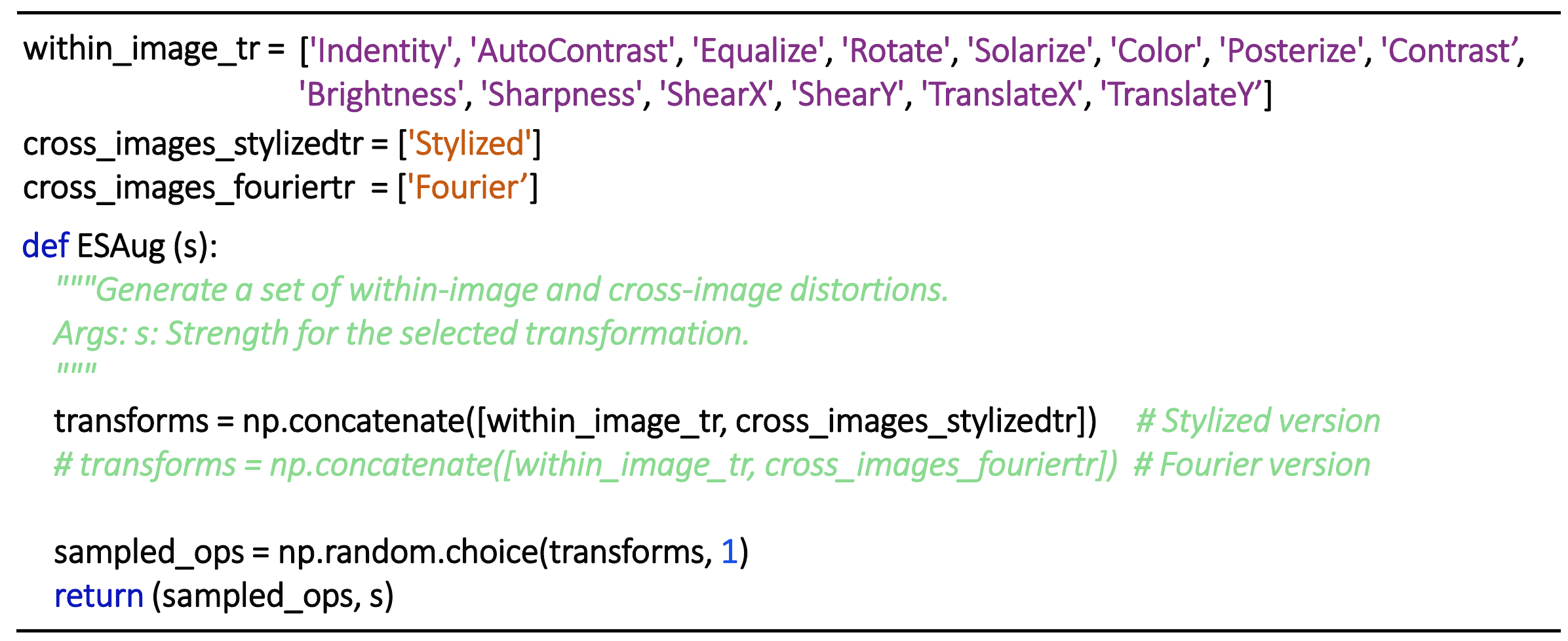}
	\caption{Python code for ESAug based on numpy.} 
	\label{fig:ESAug}
\end{figure}

\noindent \textbf{Style-based augmentation.}
The core idea of style-based augmentation is motivated by the fact that different domains contain different style information. To learn domain-invariant features, a good model should be invariant to style changes among domains. To achieve it, we include a random stylization as an augmentation to enhance the network. Specifically, we use a pre-trained AdaIN~\cite{huang2017arbitrary} which can achieve fast stylization to arbitrary styles. We randomly sample an image $x_j$ from source domains and $y_j = y_i$. Then, we randomly stylize an image $x_i$ with the style of  $x_j$ and obtain $\hat{x}_i$. 
We denote the set $\gA^s$ to represent the transformation from $x_i$ to $\hat{x}_i$.
The final augmentation set can be defined as $\gA'$ and $\gA' = \gA \cup \gA^s$.

\noindent \textbf{Singular Augmentation}
Second, we verified that the learning with replay should be performed with a  randomly selected transformation from the augmentation list with some randomness instead of sequentially applying several transformation functions. 
Prior work also observed that a cascade of successive compositions could produce images that drift far from the original image and lead to unrealistic images~\cite{hendrycks2019augmix}. Therefore, we randomly select a singular augmentation from the augmentation list. 
Due to the design of singular augmentation, we randomly set parameter $\xi$ in Fourier-based augmentation to increase some randomness to the augmentation strength, which is different from the fixed $\xi$ and a sequential augmentation in~\cite{xu2021fourier}. For style-based augmentation, we increase the augmentation randomness by randomly choosing an image from the dataset as the target to stylize image $x_i$. 
Fig.~\ref{fig:ESAug} shows Python code for ESAug based on numpy.

\begin{algorithm}[!t] 
	\caption{TripleE for Domain Generalization} 
	\label{alg1} 
	\begin{algorithmic}[1]
		\REQUIRE Source domains $\gD$, batch size $b$, replay times $r$, sub-sampling times $m$
		\REQUIRE $m$ neural networks that share a same structure $f_1(\theta),...,f_m(\theta)$
        \REQUIRE A set of exhaustive augmentations $\gA'$ 
		
		\WHILE{not converged}
		\STATE Initialize model parameters for $f_1(\theta),...,f_m(\theta)$
        
		\FOR{epochs = 1 to $N$}
		\STATE Randomly split $\gD$ into $ D^1, D^2, ..., D^m $ {\color{black}{      \# EReplayD}}
		\FOR{each sub-dataset $D^i$ in $\gD$}
	\FOR{each step}
		\STATE Sample $\gB$ from $D^i$
		\STATE Sample an augmentation $a$ from $\gA'$  { \color{black}{   \# ESAug}}
		\STATE Sample a strength $s$ from $\left \{0, ..., 30\right \}$ 
		\STATE Obtain $\gB'$ by augmenting $\gB$ with $r$ different transformations from $\gA'$, where each augmentation is $a(x,s)$  {\color{black}{   \# EReplayB}}
		\STATE Optimize $f_i(\theta)$ using Eq.~\ref{eq:2}. 
	\ENDFOR
		\ENDFOR
		\ENDFOR 
		\ENDWHILE 
	\RETURN $\theta_1, ..., \theta_m$ 
	\end{algorithmic} 
\end{algorithm}
\vspace{-7mm}

\subsection{EReplayD: Episodically Replay on Dataset}
% To achieve a good DG model, the key idea of ESub is to regard 

The idea of learning with replay is further performed on a sub-dataset, namely EReplayD. The key motivation of EReplayD is that learning with periodic updates of subsets can achieve more diverse models, leading to an improved model ensemble. 
%good DG model should be ensembled 
%perform well and not be affected by some outliers in the source domains. 
%However, it is hard to identify outliers from training images in source domains. 
%EReplayD reduces the effects of outliers on models by performing episodically replay on dataset, \ie, periodic subsampling. 
Specifically, instead of training a model on a whole training dataset $\mathcal{D}$, EReplayD randomly splits the training dataset $\mathcal{D}$ into $m$ sub-datasets $D^1, D^2,..., D^m$ and trains $m$ models on each sub-dataset. Note that the dataset will be randomly split into $m$ sub-datasets every epoch such that each model can focus on learning each sub-dataset while keeping a chance to visit the overall dataset. The final prediction will be an ensemble of predictions from $m$ models. Fig.~\ref{fig:data}(b) shows the comparison of the traditional model ensemble and our EReplayD.  
Algorithm 1 shows the overall training pipeline of TripleE.

% \begin{algorithm}
%     \caption{TripleE for Domain Generalization}
%     \SetKwFunction{isOddNumber}{isOddNumber}
%     \SetKwInOut{Require}{Require}
%     \SetKwInOut{WHILE}{while}
    
%     \Require{Source domains $\gD$}
%     \Require{Batch size $b$, replay times $r$, sub-sampling times $m$}
%     \Require{Neural networks $\left\{f_1(\theta),...,f_m(\theta) \right\}$}
%     \Require{A set of augmentations $\gA$}
    
%     \WHILE{not converged}
%     \ENDWHILE
%     $newList = [\ ]$

%     \tcc{For odd elements in the list, we add 1, and for even elements, we add 2.
%     After the loop, all elements are even.}
%     \For{$i \leftarrow 0$ \KwTo $n-1$}{
%         \eIf{$\isOddNumber(a_i)$}{

%             $newList.append(a_i + 1)$ \tcp*[f]{Some thought-provoking comment.}
%          }{
%             \tcp{Another comment}
%             $newList.append(a_i + 2)$
%          }
%     }
%     \KwRet{$newList$}
% \end{algorithm}

\if 1 
To learn domain-invariant representations across multiple source domains, we introduce domain contrast loss function.  
The key insight is to pull together representations for images with the same labels across domains (\emph{positives}) and push away representations of images with different labels (\emph{negatives}). 
Unlike contrastive learning methods that rely on large batch size to obtain sufficient positive and negative pairs, our method is tailored for DG and takes the generalization benefits of small-batch training.

\subsection{Instance Repetition}
\label{sec:1}
Our method aims to learn domain-invariant features with small-batch training, whereas small-batch training may cause inadequate samples in each feedforward calculation. 
It is problematic for a contrastive loss, which relies on multiple contrastive pairs to construct the loss function. 
Specifically, with high probability, each class may only present once in the sampled batch, making it infeasible to learn domain-invariant representations.

In this regard, we present instance repetition, which  simultaneously leverages the benefits of small-batch training and provides multiple comparable samples for domain-invariant representation learning (see Sec.~\ref{section:3.2}). 
We denote $ \mathcal{D} = \left \{ (\vx_i, y_i) \right \}_{i=1}^{N} $ as a mixed training dataset from multiple source domains, consisting of $N$ image-label pairs, where $y_i$ denotes the class label of image $\vx_i$. Note we do not assume we have access to the domain label, \ie, which source domain a given $\vx_i$ belongs to.
Our goal is to learn a feature embedding network $f_{\theta}(\cdot)$ from the training dataset.  $f_{\theta}(\cdot)$ maps the input $\vx_i$ to a low-dimensional embedding $ f_{\theta}(\vx_i) \in \mathbb{R}^{d}$, where $d$ is the dimension of the embeddings. 
% For simplicity, we denote the feature representation $f_{\theta}(x_i)$ of an image $x_i$ as $z_i$, and we assume all the features are L2 normalized, \ie, $ \left \| z_i \right \|_{2} = 1$. 
Let $\mathcal{B} \subseteq   \mathcal{D}$ denote a randomly sampled batch during training; its cardinality is the batch size.

To achieve instance repetition, assume we have a set of functions $\{T_1, T_2, \dots\}$ for data augmentation, each of which takes as input one image $\vx_i$ and outputs a perturbed image, we apply the data augmentation for each sample in $\mathcal{B}$ for $M$ times.
We denote the newly obtained sample set as $\mathcal{B}^{'}$, where $\mathcal{B}^{'} = \left \{  (T_j(\vx_i), y_i)\: |\: (\vx_i, y_i) \in \mathcal{B},\: j = 1,2,\cdots,M \right \}$ and $|\mathcal{B}^{'}| = M |\mathcal{B}|$. Naturally, a perturbed image $T_j(\vx_i)$ has the same label with the input $\vx_i$.
Then, our method will optimize the loss (detailed in Sec. \ref{section:3.2}) over this updated batch $\mathcal{B}^{'}$. With instance repetition, we can bring the benefits of small-batch training to domain generalization, which effectively avoids overfitting and guarantees that we have sufficient positive and negative pairs for learning domain-invariant representations to be introduced in the following section.

% The common training update rule for each batch is described in Eq.~(\ref{eq:1}).  
% \begin{equation}
% {\theta}_{t+1} = {\theta}_t - \alpha \frac{1}{B}  \sum_{i \in \mathcal{B}(t)}  \nabla_{\theta} \ell(T(x_i), y_i),
% \label{eq:1}
% \end{equation}
% where $T, \alpha, B$ denote the standard data augmentation and learning rate, respectively. $\mathcal{B}(t)$ denotes the set of sampled data indices at time point $t$.  $\ell$ denotes loss function defined in Sec~\ref{section:3.2} that we aim to optimize.  

% To achieve instance repetition, we introduce $M$  samples for the same input by applying random data augmentations $T$. The modified learning rule is:
% \begin{equation}
% {\bf w}_{t+1} = {\bf w}_t - \alpha \frac{1}{M \cdot B}  \sum_{i\in \mathcal{B}^{'}(t)}  \nabla_{\bf w} \ell({\bf w}_t, T(x_i), y_i),
% \end{equation}

% \begin{equation}
% {\bf w}_{t+1} = {\bf w}_t - \alpha \frac{1}{M \cdot B}  \sum_{i\in \mathcal{B}(t)} \sum_{j=1}^M  \nabla_{\bf w} \ell(T_j(x_i), y_i),
% \end{equation}\hr{how about this?}
% where sample set $\mathcal{B}(t)$ is enlarged to $\mathcal{B}^{'}(t)$, consisting of $B$ samples with $M$ different transformations. In the following, we omit $t$ for simplification. 
\vspace{-0.5mm}
\subsection{Instance-repeated Domain Contrast}
\label{section:3.2}

% Since IDC trains with small batch size and it is likely that only one image is sampled for each class in the batch.
% In this case, there are no positive pairs in the batch, and it is problematic to achieve the optimization goal. 
% As analyzed in Sec.~\ref{sec:1}, instance repetition provides more positive pairs in each feedforward evaluation, making it feasible to learn domain-invariant features by domain contrast under small-batch training. 
% $P(i) = \left. \{ p \in \mathcal{B}^{'}(i) \: | \:y_p = y_i   \right. \}$
% $\gP(i)=\{j\in[0,|\gB'|)\:|\:y_j=y_i\}$
We denote $\gP(i)=\{j\in[0,|\gB'|)\:|\:y_j=y_i,\: j\neq i\}$ as the set of indices of all positives (with the same label) in the batch $\gB'$ distinct from $i$ and $\left | \gP(i) \right |$ is its cardinality. Note that we guarantee that $\forall i, \: |\gP(i)| \geq M-1$ since we use instance repetition for the updated batch $\gB'$. 
The positive pairs in the batch include the augmented views of image $\vx_i$ and may also contain other image samples that have the same label with image $\vx_i$ from the current or different source domains.
The negative pairs in the batch are images with different labels. 
We use the cosine distance to measure the similarity of a pair of embeddings of image $\vx_i$ and $\vx_j$, denoted by $s_{i,j}$, we have
\begin{equation}
\vspace{-1mm}
s_{i,j} = \frac{f_\theta(\vx_i) \cdot f_\theta(\vx_j)}{ \left \| f_\theta(\vx_i) \right \|  \left \| f_\theta(\vx_j) \right \|}. 
\end{equation}

To optimize for domain-agnostic representation, the similarity among positive pairs should be enlarged and the similarity among negative pairs should be minimized. Accordingly, we 
define the domain contrast loss as 
\begin{equation}
\ell  = - \sum_{i}\frac{1}{ \left | \gP(i)\right |}  \sum_{j \in \gP(i)}{\rm log} (\frac{ {\rm exp} (s_{i,j} / \tau ) }{\sum_{k=1}^{M \cdot |\gB|}  \mathbbm{1}_{[k \neq i]} {\rm exp}(s_{i,k}/\tau) }  ), 
\label{eq:4}
\end{equation}
where $\tau>0$ is the temperature; $s_{i,j}, s_{i,k}$ denote the feature similarities among positive pairs and negative pairs, respectively; $M \cdot |\gB|$ is the size of the sample set $\mathcal{B}^{'}$.  
%Through minimizing Eq.~\ref{eq:4}, the embeddings from the positive pairs will be pulled together and the embeddings from the negative pairs will be pushed away. %\hr{add one sentence here to briefly conclude this subsection and say why this should work well} 
The nominator contains  all positive pairs, including the augmented samples as well as samples from other source domains with the same label. The denominator preserves the summation over negatives. The loss encourages the encoder to give closely aligned representations for \emph{all} positive pairs including those across domains. The formulation also gives a large penalty to hard negatives with high similarity or positives with low similarity, resulting in better optimization for representation learning.

Different from contrastive learning used for unsupervised representation learning, our domain contrast loss contains positive pairs from different groups. One group is the self-augmented samples and the other group is the distinct images with the same label.  
We introduce a hyperparameter $\beta$ to weight the loss of different positive pairs, 
and Eq.~\ref{eq:4} is modified to Eq.~\ref{eq:new_2}. 
\begin{equation}
\begin{aligned}
\ell  = - \sum_{i}\frac{1}{ \left | \gP(i)\right |} \bigg( \sum_{j \in \gP_a(i)} {\rm log} (\frac{ {\rm exp} (s_{i,j} / \tau ) }{\sum_{k=1}^{M \cdot |\gB|}  \mathbbm{1}_{[k \neq i]} {\rm exp}(s_{i,k}/\tau) }  ) \\   +  \beta \sum_{j \in \gP_d(i)} {\rm log} (\frac{ {\rm exp} (s_{i,j} / \tau ) }{\sum_{k=1}^{M \cdot |\gB|}  \mathbbm{1}_{[k \neq i]} {\rm exp}(s_{i,k}/\tau) })\bigg), 
\label{eq:new_2}
\end{aligned}
\end{equation}
% where $\beta$ is the ratio for (distinct images with the same labels) /(self-augmented positive pairs). 
where $\gP_a(i)$ denotes the set of datapoints that are the augmented instance of $i$; $\gP_d(i)$ denotes the distinct images with the same label as $i$; and $\gP(i)=\gP_a(i) \cup \gP_d(i)$.  
Hence, through minimizing Eq.~\ref{eq:new_2}, we can get a robust category-specific domain-invariant representation.

\subsection{Embedding Matching Scheme}\label{sec:matching} 
After training our model using instance-repeated domain contrast, we achieve a latent embedding space that preserves the high-level category-aware representations irrespective of the domain information. Our framework is very flexible and we propose two options to perform downstream predictions for data in the target domain using the structured embeddings in the vector space.  

% Our framework is very flexible and we present two options to obtain the final prediction of the label for data in the target domain. 
Since the structured embedding space is already informative enough and clustered according to the categories across domains, we can perform feature matching in the embedding space in a completely unsupervised manner.
One direct way is to design a linear classifier to map the embedding space to the class probability, which can be trained using $(\vx,y)$ pairs from the source domains. Note that we fix the weights of the embedding network $f_\theta$ in this process and only train the classifier. We further present an alternative that saves the effort of training additional parameters. 
Specifically, given the embedding of a target and all the data from the source domains, we employ a simple nearest neighbour algorithm to retrieve $k$-nearest neighbours for a target embedding in the vector space. Then we make the final prediction using the most frequent labels of the $k$ neighbours, which are labeled images from the source domains. In this case, we do not need to train additional parameters and it can be fairly efficient in domain generalization tasks where the data from the source domain are scarce. 
% We can also use several approximate nearest neighbour algorithms~\cite{guo2020accelerating,johnson2019billion} to scale to larger datasets, which is out of the scope of this paper.

\subsection{Model Details}

\noindent {\bf Network Architecture.}
We employ the standard ResNet-18 and ResNet-50~\cite{he2016deep} as the network backbones for $f_\theta$. We further replace the final classification layer in the original ResNet~\cite{he2016deep} with a fully-connected layer with 128 output channels to calculate the embeddings for the input image. We follow exactly the same protocol as in prior work~\cite{wang2020learning,huang2020self, Carlucci_2019_CVPR} for fair comparison and initialize the weights of $f_\theta$ with the ImageNet pre-trained weights.
% The ImageNet pre-trained weights are employed as the network initialization, following the same setting in prior works. 

\noindent {\bf Augmentation.}
We adopt several standard strategies, including (1) we randomly cropped the images to retain between 80\% to 100\%; (2) we randomly applied horizontal flipping and gray scaling with 0.5 and 0.1  probability, respectively; (3) we also randomly changed the brightness, contrast, and saturation of the input image with a random factor chosen uniformly from [0.6, 1.4]. 

\noindent {\bf Training.}
We also adopt the exact same training protocol as the prior methods~\cite{wang2020learning, Carlucci_2019_CVPR, huang2020self}.
Specifically, IDC was trained with the SGD optimizer~\cite{bottou2010large}.  
The learning rate was set to 0.001 and annealed to 0.0001 after 80\% of the training epochs.
We train the model for 100 epochs in each run, and select the best model using the validation set.
For the configurations of IDC, the temperature $\tau$ was set as 0.07, following the settings in~\cite{chen2020simple}. The batch size $B=|\gB|$ was 4, and $M$ was 4. We also analyze the effects of different $B$ and $M$ in ablation studies (see Sec.~\ref{sec:ablation}).
% \noindent {\bf Inference stage}
% Given a new datapoint in the target domain, we first embed it using a trained embedding network $f_\theta$ and then perform matching in the embedding space by using an additional classifier or $k$-nearest neighbor (which we set $k=500$) as discussed in Sec. \ref{sec:matching}. 

%\hr{can remove this paragraph if we are short of space. If removed, also need to change the title}
% Given a trained model, we compare feature embeddings of target data points with the source data points. 
% We empirically set $k$ as 500 and the effects of different $k$ for generalization performance are analyzed in Sec~\ref{sec:para}. 

\fi 

%% file: sections/5.Experiments.tex
\section{Experiments}

\subsection{Datasets and Settings}

\noindent \textbf{Datasets.} We evaluate our method on six widely-used datasets in domain generalization, \ie, \textbf{Digits-DG}, \textbf{PACS} and \textbf{Office-Home}, \textbf{VLCS}, \textbf{TerraInc} and \textbf{DomainNet}.  
Digits-DG~\cite{zhou2020learning} contains four different digit datasets, \ie, MNIST~\cite{lecun1998gradient}, MNIST-M~\cite{ganin2015unsupervised}, SVHN~\cite{netzer2011reading}, and SYN~\cite{ganin2015unsupervised}. 
PACS~\cite{li2017deeper} contains 9,991 images of four domains (Artpaint, Cartoon, Sketches, and Photo) and 7 classes. Office-Home~\cite{venkateswara2017deep} contains 15,558 images of 65 classes, distributed across 4 domains (Artistic, Clipart, Product, and Real world). VLCS~\cite{fang2013unbiased} includes photographic domains (Caltech101, LabelMe, SUN09, VOC2007) with 10,729 images and 5 classes.
TerraInc dataset~\cite{beery2018recognition} contains 24,788 images with 10 classes over 4 domains. 
DomainNet dataset~\cite{peng2019moment} has 586,575 images with 345 classes over 6 domains.

\noindent \textbf{Evaluation protocol.} Following prior works~\cite{xu2021fourier,Carlucci_2019_CVPR}, we train our model on the training splits and select the best model on the validation splits of all source domains. For testing, we evaluate the selected model on all images of the held-out target domain. For performance evaluation, we report top-1 classification accuracy on each test domain and average accuracy accordingly. 

\subsection {Implementation Details}  
Here we briefly introduce the main details for training our method. 

\noindent \textbf{Basic details:} 
All experiments were implemented in PyTorch~\cite{NEURIPS2019_9015} and run on a machine with an NVIDIA RTX 3090 GPU. We trained all the models with the cross-entropy loss and an InfoNCE loss function~\cite{khosla2020supervised}. 
For Digits-DG, we use the same backbone network as~\cite{zhou2020learning,Carlucci_2019_CVPR}. We train the network from scratch using SGD and the initial learning rate is set to 0.01. For PACS and Office-Home, we use ImageNet pretrained ResNet~\cite{he2016deep} as our backbone. The initial learning rate of PACS is 0.0001 and decayed by 0.5 every 30 epochs. The initial learning rate of Office-Home is 0.01
and decayed by 0.5 every 30 epochs.
The network is totally trained for 100 epochs as \cite{wang2020learning}. For VLCS and TerraInc, we use the ImageNet pretrained ResNet-50 ~\cite{he2016deep} as our backbone. The initial learning rate is 0.001 and decayed by 0.5 every 40 epochs. We trained 100 epochs for these two datasets. For DomainNet, we also use the ImageNet pretrained ResNet-50~\cite{he2016deep} as our backbone. The initial learning rate is 0.001 and decayed by 0.5 every 16 epochs. We trained 40 epochs for this dataset.

\noindent \textbf{Method-specific details:} For Digits-DG and PACS, both batch size $b$ and repeat time $r$ are set to 4. For Office-Home, both $b$ and $r$ are set to 32. For VLCS, TerraInc, and DomainNet, the batch size $b$ is 16, and the repeat time $r$ is 4. For all experiments, we set the sampling times $m$ to 3.

\begin{table}[!t]
	\centering
	\caption{Domain generalization results on the Digits-DG dataset (Backbone: 4 conv layer, which is used in~\cite{zhou2020learning}). ``DL'' refers to using domain label or not.}
	\resizebox{\linewidth}{!}{{\begin{tabular}{c|c|cccc|c}	\toprule	
	& DL & SYN & SVHN & MNIST-M & MNIST & Avg. 			\tabularnewline \hline 	   Vanilla~\cite{zhou2020deep} & \ding{55} & 78.6 & 61.7 & 58.8 & 95.8 & 73.7 
                \tabularnewline 
           \hline 
                CCSA~\cite{motiian2017unified} & \ding{51} & 79.1 & 65.5 & 58.2 & 95.2 & 74.5 
                \tabularnewline 
                MMD-AAE~\cite{li2018domain} & \ding{51} & 78.4 & 65.0 & 58.4 & 96.5 & 74.6 
                \tabularnewline 
                 CrossGrad~\cite{shankar2018generalizing} & \ding{51} & 80.2 & 65.3 & 61.1 & 96.7 & 75.8 
                \tabularnewline

                % RSC~\cite{wang2020learning}  & & 86.1 &	67.7 &	53.1 & 	94.9 &	 75.4
                % \tabularnewline 

                DDAIG~\cite{zhou2020deep} & \ding{51} & 81.0 & 68.6 & 64.1 & 96.6 & 77.6
                \tabularnewline 
              
                 L2A-OT~\cite{zhou2020learning} & \ding{51} & 83.2 & 68.6 & 63.9 & 96.7 & 78.1 
                \tabularnewline 
                
            	STEAM~\cite{chen2021style} & \ding{51} & 92.2 & 76.0 & 67.5 & 96.8 & 83.1
                \tabularnewline 
                  \hline 
               JiGen~\cite{Carlucci_2019_CVPR} & \ding{55} & 74.0 & 63.7 & 61.4 & 96.5 & 73.9 \tabularnewline 
             SFA~\cite{li2021simple} & \ding{55} & 85.0 & 70.3 & 66.5 & 96.5 & 79.6
                \tabularnewline 
                
               FACT~\cite{xu2021fourier} & \ding{55} & 90.3 & 72.4 & 65.6  & 97.9 & 81.5
                \tabularnewline 
                
            \hline 
			TripleE-Style {\bf (ours)}  & \ding{55} & 95.33 & \textbf{81.75}  &  72.52 & 98.28 & 86.97	
			\tabularnewline 
				
			TripleE-Fourier {\bf (ours)}  & \ding{55} & \textbf{95.48} & 81.22  &  \textbf{73.15} & \textbf{98.33} & \textbf{87.04}	
							
				\tabularnewline
				
				\bottomrule
		\end{tabular}}	\label{tab:digits}}
\end{table}

\subsection{Evaluation on six DG benchmarks}

\noindent \textbf{Digits-DG dataset.} Table~\ref{tab:digits} shows the results on the Digits-DG dataset. 
It is clear that our TripleE method outperforms the prior state-of-the-art DG methods STEAM~\cite{chen2021style} and FACT~\cite{xu2021fourier} by an average of 3.9\% and 5.5\% on top-1 classification accuracy, respectively.  
Please note that STEAM~\cite{chen2021style} not only requires domain label information but also requires carefully designing a teacher-student model and multiple memory banks to store the style features for each source domain. 
In contrast, our TripleE method does not require a domain label and does not need any particular architecture design.  
Using simple data augmentation and sub-sampling ensemble, our method can achieve the best performance. 
Notably, FACT~\cite{xu2021fourier}
is a dual-model (teacher-student model) that relies on Fourier transforms to exchange domain information.
However, our TripleE model obtains significant improvements (+8.82\%
and +7.55\% on SVHN and MNIST-M, respectively) over FACT~\cite{xu2021fourier}.
The result demonstrates that our simple approach can effectively deal with large domain shifts caused by complex backgrounds and cluttered digits.

\begin{table}[!t]
	\centering
	\caption{Domain generalization results on the PACS dataset. ``S,C,A,P'' refer to ``Sketch, Cartoon, Art painting, and Photo'', respectively. Each reported result of our method is averaged over three runs. ~\dag use \emph{test-domain-validation-set}, where some target images are used to select the best model. ``DL'' refers to using domain label or not.
	}
	\resizebox{\linewidth}{!}{{\begin{tabular}{c|c|cccc|c}
			\toprule[1.5pt]
				& DL & S & C & A &  P &   Avg. 
			\tabularnewline \hline 	
			
            \multicolumn{7}{c}{ResNet-18} 
			\tabularnewline \hline 
			Vanilla & \ding{55} &  69.64 & 75.65 & 77.38 &  94.25  & 79.23 
			\tabularnewline 			
			
			MMD-AAE~\cite{li2018domain}  &\ding{51} &   64.2 & 72.7 & 75.2 & 96.0  & 77.0
			\tabularnewline 
			
			CCSA~\cite{motiian2017unified}  &\ding{51}  &  66.8 & 76.9 & 80.5 & 93.6  & 79.4
			\tabularnewline 
		
			D-SAM~\cite{d2018domain}  & \ding{51}  &  77.83 & 72.43 & 77.33 & 95.30 &  80.72
			\tabularnewline 
			
			MASF~\cite{dou2019domain} &\ding{51} &  71.69 & 77.17 & 80.29 & 94.99  & 81.04
			\tabularnewline 	
			
			DMG~\cite{chattopadhyay2020learning}  & \ding{51} & 75.21 & 80.38  & 76.90 & 93.35  & 81.46
			\tabularnewline 
			
			Epi-FCR~\cite{Li_2019_ICCV} &\ding{51} &  73.0 & 77.0 & 82.1  & 93.9  & 81.5
			\tabularnewline	
			CuMix~\cite{mancini2020towards} & \ding{51} & 72.6 & 76.5  & 82.3 & 95.1 & 81.6
			\tabularnewline			
			
			MetaReg~\cite{NIPS2018_7378} &\ding{51} &  70.3  & 77.2  & 83.7   & 95.5  & 81.7
		\tabularnewline

			L2A-OT~\cite{zhou2020learning}  & \ding{51} & 73.6 & 78.2  & 83.3 & 96.2 & 82.8
			\tabularnewline
			
			SelReg~\cite{kim2021selfreg}
			& \ding{51} & 77.47  & 78.43  & 82.34   & 96.22  & 83.62
			\tabularnewline	
			STEAM~\cite{chen2021style}   & \ding{51} & 82.9 & 80.6  & 85.5  & 97.5  & 86.6
			\tabularnewline

			\hline

			JiGen~\cite{Carlucci_2019_CVPR}  &\ding{55}  &  71.35 & 75.25 & 79.42 & 96.03  & 80.51
			\tabularnewline 	
						
			SFA~\cite{li2021simple}  & \ding{55}  & 73.7 & 77.8  & 81.2  & 93.9 & 81.7
			\tabularnewline
			
			MMLD~\cite{matsuura2020domain} & \ding{55} & 72.29  & 77.16 & 81.28 & 96.09& 81.83 \tabularnewline

			EISNet~\cite{wang2020learning} & \ding{55} & 74.33  & 76.44    & 81.89  & 95.93  & 82.15
			\tabularnewline
			
			InfoDrop~\cite{shi2020informative}  & \ding{55}
			&   76.38 & 76.54 & 80.27 & 96.11 & 82.33   
			\tabularnewline

			pAdaIN~\cite{nuriel2021permuted} & \ding{55} & 75.13 & 76.91  & 81.74 & \textbf{96.29} & 82.51
			\tabularnewline

			FACT~\cite{xu2021fourier}  &  \ding{55}   & 79.15 & 78.38  & \textbf{85.37}  & 95.15 & 84.51
			\tabularnewline

% 			LTPT~\cite{pandey2021generalization}  &  &  81.79 & \textbf{81.26}  & \textbf{86.39}  & 97.15  & 86.65
% 			\tabularnewline
			
	        \hline 
			TripleE-Style {\bf (ours)}  & \ding{55}  &\textbf{84.53}  &\textbf{81.02}  &85.16   &96.23   &\textbf{86.74}
			\tabularnewline
											
			TripleE-Fourier {\bf (ours)}  & \ding{55}  &84.45  &80.84  &85.3   &96.27   &86.72 
			\tabularnewline

			\hline 
			\multicolumn{7}{c}{ResNet-50} \\
			
			\hline 
			Vanilla  & \ding{55} & 69.69 & 78.61 & 81.47 & 94.83 & 81.15  
			\tabularnewline	
			
			MASF~\cite{dou2019domain}  &  \ding{51}  &   72.29 &  80.49 & 82.89  & 95.01  & 82.67  
			\tabularnewline	
			
			MetaReg~\cite{NIPS2018_7378} &  \ding{51} & 70.30 & 79.20  & 87.20 & 97.60 & 83.60
			\tabularnewline 
			
			DMG~\cite{chattopadhyay2020learning} &  \ding{51} & 78.32 & 78.11  & 82.57 & 94.49  & 83.37
			\tabularnewline 
			\hline 
			
			pAdaIN~\cite{nuriel2021permuted} & \ding{55} & 77.37 & 81.06  & 85.82 & 97.17 & 85.36
			\tabularnewline
			
			EISNet~\cite{wang2020learning}    &  \ding{55} &  78.07 & 81.53  & 86.64  & 97.11   & 85.84 
			\tabularnewline	
			
			Contrastive ACE~\cite{wang2021contrastive}~\dag & \ding{55}  &   80.6 & 81.9 & 88.8  & 97.7  & 87.3
			\tabularnewline

		SWAD~\cite{cha2021swad} & \ding{55} &  -  & -   & -  &  -   & 88.1  
			\tabularnewline
			
% 		AdaClust~\cite{thomas2021adaptive} & &  -  & -   & -  &  -   & 89.2 
% 			\tabularnewline
		
		FACT~\cite{xu2021fourier}  &   \ding{55}  & 84.46 & 81.77  & 89.63  & 96.75 & 88.15
	\tabularnewline
		
% 		LTPT~\cite{pandey2021generalization}  & &  86.20 & 85.19  & 90.25 & 98.97  & 90.15
% 			\tabularnewline
			\hline  
		
		TripleE-Style {\bf (ours)}  & \ding{55}  &\textbf{88.19}  &85.2  &90.97   &97.72 & \textbf{90.52}
			\tabularnewline

			TripleE-Fourier {\bf (ours)}  & \ding{55}  & 86.23  & \textbf{85.37}  &\textbf{91.06}   &\textbf{98.32}   & 90.25
			\tabularnewline

			\bottomrule[1.5pt]
		\end{tabular}}
		\label{tab:PACS_res18}}
	
\end{table}

\noindent \textbf{PACS dataset.} Table~\ref{tab:PACS_res18} shows the results on the PACS dataset. We can see that our TripleE can clearly outperform the prior best DG method under the same setting by 2.2\% and 2.4\% using ResNet-18 and ResNet-50, respectively.
Notably, TripleE-Style and TripleE-Fourier can reach comparable performance on both PACS and Digits-DG, showing that these two kinds of augmentation can encourage the network to learn high-level semantics by exchanging low-level statistics among images. 
Note that FACT~\cite{xu2021fourier} built a dual-model (teacher-student models) and encourages high-level semantic exchanges among domains to improve generalization.  
Unlike this method, our TripleE is a simple 
augmentation and ensemble method. It is worth mentioning that SFA~\cite{li2021simple} is also an augmentation-based method that augments feature embeddings to improve domain generalization. However, in terms of accuracy, our method can excel all other DG methods and reach a comparable performance with STEAM~\cite{chen2021style}, which uses domain labels and builds a complex network with at least three memory banks.

\if 1 
Our SFAA improves the simple variant SFA-S with 0.7\% accuracy,
resulting in a clear improvement margin of 2.6\% accuracy over the vanilla ERM method and outperforming the data augmentation DG method CrossGrad [33] with 3.9\%. Although not as good as the recent state-of-the-art methods, such
as EISNet and L2A-OT [45], our method wins in term of its efficiency and simpleness. And note that L2A-OT uses domain labels at training, whereas we do not
\fi

% % shows the results with ResNet-18 and ResNet-50 backbones~\cite{he2016deep}. 
% Different from the other methods, we present a novel paradigm to improve domain generalization. 
% We find that our IDC method achieves the best performance on \emph{Sketch} and \emph{Art Painting}, outperforming the previous best result by around 5.8\% and 2.4\% accuracy with ResNet-18. 
% Similarly, our method consistently achieves the best performance on \emph{Sketch}, \emph{Cartoon} and \emph{Art Painting} with the ResNet-50 backbone.  
% Across all domains, we can observe IDC reaches an average accuracy of 85.78\% with ResNet-18 and 88.09\% with ResNet-50, setting a new state-of-the-art result on this benchmark dataset.  
% Notably, one state-of-the-art DG method~\cite{huang2020self} reported 85.15\% and  87.83\% accuracy with ResNet-18 and ResNet-50. However, the results are hard to reproduce using the official code at the current stage, as shown in GitHub\footnote{https://github.com/DeLightCMU/RSC/issues/} and Table 3 in~\cite{nuriel2020permuted}. 

\begin{table}[!t]
	\centering
	\caption{Domain generalization results on the Office-Home dataset. ``DL'' refers to using domain label or not (Backbone: resnet-18). }
	\resizebox{\linewidth}{!}{{\begin{tabular}{c|c|cccc|c}	\toprule
	& DL & Art & Clipart & Product & Real World & Avg. 			
	            \tabularnewline \hline 	   
	            Vanilla~\cite{zhou2020deep} & \ding{55} & 58.9 & 49.4 & 74.3 & 76.2 & 64.7
                \tabularnewline \hline

                MMD-AAE~\cite{zhou2020learning} &\ding{51} & 56.5 & 47.3 & 72.1 & 74.8 & 62.7
                \tabularnewline 
                
                CrossGrad~\cite{shankar2018generalizing} & \ding{51} & 58.4 & 49.4 & 73.9 & 75.8 & 64.4
                \tabularnewline 
               CCSA~\cite{motiian2017unified} & \ding{51} & 59.9 & 49.9 & 74.1 & 75.7 & 64.9 
                \tabularnewline 
               DDAIG~\cite{motiian2017unified} & \ding{51} & 59.2 & 52.3 & 74.6 & 76.0 & 65.5 
                \tabularnewline 
               
                L2A-OT~\cite{zhou2020learning} &\ding{51} & 60.6 & 50.1 & 74.8 & 77.0 & 65.6
                \tabularnewline 
				STEAM~\cite{chen2021style} & \ding{51} & 62.1 & 52.3 & 75.4 & 77.5 & 66.8
                \tabularnewline \hline 
                 JiGen~\cite{Carlucci_2019_CVPR} & \ding{55} & 53.0 & 47.5 & 71.5 & 72.8 & 61.2 
                \tabularnewline 
			  
			  RSC~\cite{huang2020self} & \ding{55} & 74.54 & 71.63 & 47.90 & 58.42 & 63.12 
                \tabularnewline 
			   FACT~\cite{xu2021fourier} & \ding{55} & 60.34 & 54.85 & 74.48 & 76.55 & 66.56
                \tabularnewline 
			   
		    \hline 
		    
	    	TripleE-Style {\bf (ours)}  & \ding{55} & 63.10 &  56.47 & 76.10  & 76.42 & 68.02 \tabularnewline
	    					
			TripleE-Fourier {\bf (ours)}  & \ding{55} & 62.59 & 57.55 & 75.69 & 77.48 & \textbf{68.33}
				
				\tabularnewline	\bottomrule[1.5pt]
		\end{tabular}}	\label{tab:office}}
\end{table}

\noindent \textbf{Office-Home dataset.} 
The results are shown in Table~\ref{tab:office}. Notably, a simple vanilla model shows strong results on this benchmark. Most baselines provide only marginal improvements to the vanilla model with less than 1.0\% improvements. One reason is that Office-Home is a relatively large composition of data, compared with PACS and Digits-DG, thus offering inherently bigger domain diversity in training data already~\cite{zhou2020learning}. Overall, our TripleE can surpass the best DG method FACT~\cite{xu2021fourier} and STEAM~\cite{chen2021style} by 1.77\% and 1.53\%, respectively. The state-of-the-art result on Office-Home dataset further validates the effectiveness of our proposed EReplayB, ESAug, and EReplayD.

\noindent \textbf{VLCS, TerraInc and DomainNet datasets.} Table~\ref{tab:multiple} shows the comparison with existing state-of-the-arts on multiple DG benchmarks. 
SWAD~\cite{cha2021swad} is one of the existing DG methods. In all experiments, our TripleE achieves significant performance gain against the previous best method SWAD~\cite{cha2021swad}. The comparison shows that our TripleE could serve as a stepping stone for future research in domain generalization.  

%Fishr~\cite{rame2022fishr} ignored XXX. 

\begin{table}[htbp]
	\centering
	\caption{Comparisons of results on PACS, VLCS, OfficeHome, TerraInc, and DomainNet. Except for our method, other results are adopted from \cite{cha2021swad}. 
	%\xmli{To HF: Complete the results in this Table.}
	}
	\resizebox{\linewidth}{!}{{\begin{tabular}{c|c|ccccc}	\toprule	
	& Avg. & PACS & VLCS & OfficeHome & TerraInc & DomainNet    		\tabularnewline
    \hline 
    Mixstyle~\cite{zhou2021domain} & 60.3 & 85.2 & 77.9 & 60.4  & 44.0  &  34.0
    \tabularnewline
     
     C-DANN~\cite{li2018deep} & 62.6 & 82.8 & 78.2 & 65.6 & 47.6 & 38.9   
    \tabularnewline
    
     DANN~\cite{ganin2016domain} & 63.1& 84.6 & 78.7 & 65.4 & 48.4 & 38.4   
    \tabularnewline 
    
    ERM~\cite{gulrajani2020search} & 63.8 & 85.7 & 77.4 & 67.5 & 47.2 & 41.2   
    \tabularnewline

    Fish~\cite{shi2021gradient} & 63.9 & 85.5 & 77.8 & 68.6 & 45.1  & 42.7
    \tabularnewline
    CORAL~\cite{sun2016deep} & 64.1 & 86.0 & 77.7 &  68.6 & 46.4 & 41.8 
    \tabularnewline 
    
    MIRO~\cite{cha2022domain} & 65.9 & 85.4 & 79.0 &  70.5 & 50.4 & 44.3
     \tabularnewline
    
     SWAD~\cite{cha2021swad} &  66.9 & 88.1 & 79.1  &  70.6 &  50.0 & 46.5
    \tabularnewline \hline 
    
    %  TripleE* & - & - & 78.0 & - & 50.7 & 46.5
    %   \tabularnewline 
      
    TripleE (ours) & \textbf{69.1} & \textbf{90.5} & \textbf{80.1} & \textbf{74.0} & \textbf{52.7} & \textbf{48.2}
	\tabularnewline
	\bottomrule \end{tabular}}}	
	% \begin{tablenotes}
	% \end{tablenotes}
	\label{tab:multiple}
\end{table}

\subsection{Ablation Studies}\label{sec:ablation}

\begin{table}[!t]
	\centering
	\caption{Effectiveness of different components on Digits-DG, PACS, and Office-Home datasets (Backbone: ResNet-18). Vanilla refers to baseline used in early DG methods~\cite{zhou2020deep,Carlucci_2019_CVPR}. 
	Compared to Vanilla, recent DG methods~\cite{wang2020learning,xu2021fourier} use Baseline$\dag$
    that additionally employs color jittering as the basic data augmentation.  Baseline(ours) keeps consistent with Baseline$\dag$ except for an additional contrastive loss~\cite{khosla2020supervised}.}
    \resizebox{\linewidth}{!}{{\begin{tabular}{c|ccc|cccc|c}	\toprule[1.5pt]	
	
	\hline 
\multicolumn{9}{c}{Digits-DG dataset} 
\tabularnewline 
\hline 

	& EReplayB & ESAug  & EReplayD & SYN & SVHN & MNIST-M & MNIST & Avg. 			\tabularnewline \hline

	Vanilla~\cite{Carlucci_2019_CVPR,zhou2020deep} & - & - &  - &78.6  &61.7 &58.8 &95.8 &73.7  		\tabularnewline
					
	Baseline$\dag$~\cite{wang2020learning,xu2021fourier} & - & - &  - &89.93   &64.43  &58.32 &95.8  &77.12  		\tabularnewline
    					
    Baseline (ours) & - & - &  - & 90.63  &66.58 &58.68 &95.55 &77.86 			\tabularnewline
	
	\hline 	 Model a   &  $\checkmark$ & - & - &92.8 &75.4 &63.1 &96.4 & 81.9
                \tabularnewline 
            								
            Model b & -  & $\checkmark$   & -  &93.65  &74.37  &63.03  &97.43  &82.12
			\tabularnewline 
										
			Model c  & - & - & $\checkmark$ &90.07 &69.02 &65.47 &97.83 & 80.60	
				\tabularnewline
		      
			Model d & $\checkmark$  & $\checkmark$   & -  &94.62 &79.83 &69.85 &97.43 &85.43
			\tabularnewline 
		    				
			Model e & -  & $\checkmark$   & $\checkmark$ &94.12 &76.70 &72.05 &98.18 &85.26
			
			\tabularnewline	
			
			Model f &  $\checkmark$ & -   & $\checkmark$ &93.50  &77.07 &64.97 &97.30 & 83.21
			\tabularnewline	
			
				\hline
			TripleE (\textbf{ours})  &  $\checkmark$  &  $\checkmark$  & $\checkmark$  & \textbf{95.33}  & \textbf{81.75} & \textbf{72.52} & \textbf{98.28} & \textbf{86.97}
				\tabularnewline	
				
				\hline 
					\hline 
				\multicolumn{9}{c}{PACS dataset} 
\tabularnewline
\hline

	\hline 
	& EReplayB & ESAug  & EReplayD & S & C & A & P & Avg. 			\tabularnewline 
	\hline 	
	
\hline 
    Vanilla~\cite{Carlucci_2019_CVPR,zhou2020deep} & - & - &  -       & 69.64 & 75.65 & 77.38  & 94.25 & 79.23  
    \tabularnewline

    Baseline$\dag$~\cite{wang2020learning,xu2021fourier} & - & - &  -        & 79.74 &  79.26 & 79.49 & 93.41 & 82.98
    \tabularnewline
    				
    Baseline (ours) & - & - &  -        &76.34  &79.35  &80.96  &95.63  &83.07	
    \tabularnewline
	
	\hline 	 
	Model a   &  $\checkmark$ & - & -   & 82.78 & 78.28 & 84.57 & 95.93 &85.39
    \tabularnewline 
     				      								
	Model b & $\checkmark$  & $\checkmark$  & -  & 84.06 & 80.84 & 83.5 & 95.33 &85.93
	\tabularnewline 

	\hline
	TripleE (\textbf{ours})  &  $\checkmark$  &  $\checkmark$  & $\checkmark$  & \textbf{84.53}  & \textbf{81.02} & \textbf{85.16} & \textbf{96.23} & \textbf{86.74}
	\tabularnewline

				\hline 
					\hline 
\multicolumn{9}{c}{Office-Home dataset} 
\tabularnewline
\hline 

\hline 
	& EReplayB & ESAug  & EReplayD & Art & Clipart & Product & Real World & Avg. 	
	\tabularnewline 

\hline 
  Vanilla~\cite{Carlucci_2019_CVPR,zhou2020deep} & - & - &  -       & 58.9 &  49.4 & 74.3 & 76.2 & 64.7  
    \tabularnewline

    Baseline$\dag$~\cite{xu2021fourier} & - & - &  -        & 55.34 & 52.46 & 73.01  &  73.17 & 63.50  
    \tabularnewline

    Baseline (ours) & - & - &  - & 58.22 & 54.64 & 73.62 & 73.49 & 64.99
    \tabularnewline
    % \hline 
    %   FACT~\cite{xu2021fourier} & && & 60.34 & 54.85 & 74.48 & 76.55 & 66.56
    % \tabularnewline 

	\hline 	 
	Model a   &  $\checkmark$ & - & - & 60.73 & 53.06 & 73.80 & 75.42 & 65.75 
    \tabularnewline 
		
	Model b & $\checkmark$  & $\checkmark$   & -  & 59.87 & \textbf{58.10}  &  75.02 &  75.69 & 67.17 
	\tabularnewline 

	\hline
	TripleE (\textbf{ours})  &  $\checkmark$  &  $\checkmark$  & $\checkmark$ &  \textbf{62.59} & 57.55 & \textbf{75.69} & \textbf{77.48} & \textbf{68.33}
	\tabularnewline	
	\hline

	\bottomrule[1.5pt]
	\end{tabular}}	\label{tab:ablation_three}}
\end{table}

\noindent \textbf{Ablation of our baseline model.} We first point out that there are several inconsistent baselines in DG that share the same network structures but employ different augmentations.
As shown in Table~\ref{tab:ablation_three}, 
early DG methods~\cite{zhou2020deep,Carlucci_2019_CVPR} use random crop, and random horizontal flip as the augmentation denoted as \textbf{Vanilla}. Whereas, recent DG methods~\cite{wang2020learning,xu2021fourier} use a stronger \textbf{Baseline}$\dag$ that additionally employs color jittering as the basic data augmentation, which can enhance the performance from 73.7\% to 77.1\%. \textbf{Baseline}(ours) keeps consistent with Baseline$\dag$ except for an additional contrastive loss~\cite{khosla2020supervised}. To use the contrastive loss, we add a projection head, which sequentially consists of a fully connected layer, a batch normalization layer, a ReLu, and a fully connected layer to map the feature dimension to 128. Although with a slightly stronger baseline, we verify that the performance improvements are from our proposed TripleE instead of the baseline.

\noindent \textbf{Impact of different components.} 
Table~\ref{tab:ablation_three} shows the effectiveness of our proposed TripleE (EReplayB, ESAug, and EReplayD). 
Note that all experiments used Baseline(ours) as the backbone model. 
From experiments on Digits-DG, we can see that EReplayB, ESAug, and EReplayD can enhance our baseline by 4.0\%, 4.3\%, 2.74\%, respectively. 
From Model d and Model e, we can see that combing ESAug with  EReplayB or EReplayD can largely improve the performance to 85.43\% and 85.26\%.
With all these three components, our TripleE can achieve 86.97\%. 
The result \textbf{\textit{verifies our idea that ESAug can improve generalization performance by enlarging the data space in replays.}} 
Note that except for Model c, all other combinations of our method can excel the current best DG method FACT~\cite{xu2021fourier} (81.5\%). 
For experiments on PACS and Office-Home, we can see that training with EReplayB can outperform our baseline by $2.3\%$ and $1.7\%$ on PACS and Office-Home, respectively. Note that compared to our baseline, EReplayB only changes the batch size $b$ and repeat times $r$, showing that training with subsets can effectively improve the model generalization. By comparing \textbf{Model a} and \textbf{Model b}, we can see that using ESAug can increase the performance by $0.7\%$ and $1.4\%$ on PACS and Office-Home, respectively. These results demonstrated the effectiveness of enlarging data space when training with replays. 
Finally, we can see that using all three components can achieve the best performance on three datasets.

\begin{figure}[t]
	\centering
	\includegraphics[width=0.5\textwidth]{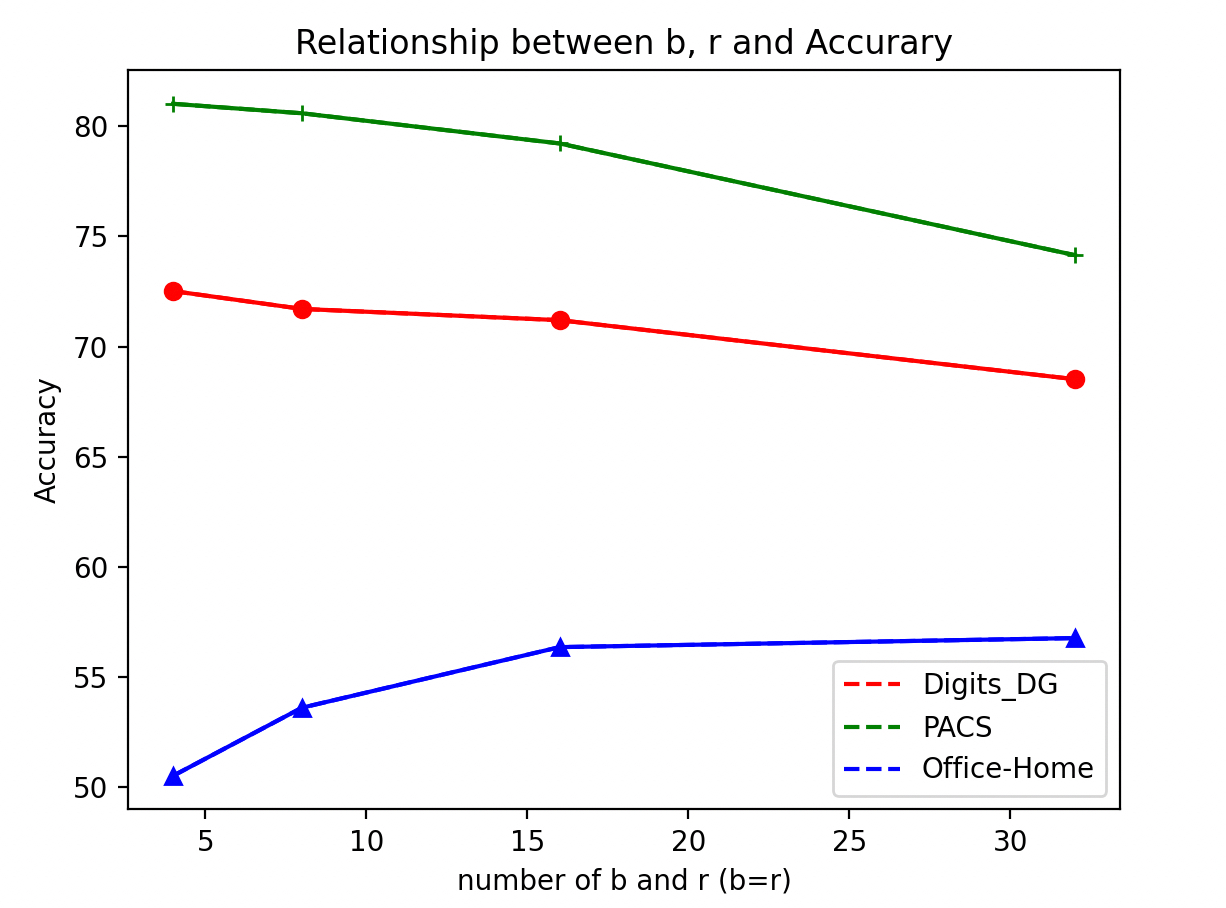}
	\caption{Effects of different $b$ and $r$ in EReplayB on three datasets. The result is reported on \emph{MNIST-M}, \emph{Cartoon} and \emph{Clipart} in Digits-DG, PACS and Office-Home, respectively.} 
	\label{fig:ereplay}
\end{figure}

\begin{table}[!t]
	\centering
	\caption{Ablation studies of ESAug on Digits-DG. $\mathcal{A}$ refers to augmentations used in~\cite{cubuk2020randaugment} and $\mathcal{A}'$ refers to include one cross-image augmentation. ``S'' refers to randomly selecting a singular augmentation, and ``Non-S'' refers to sequentially applying augmentations.}
	\resizebox{\linewidth}{!}{{\begin{tabular}{c|cc|cccc|c}	\toprule	
	
	 & AugType & AugList & SYN & SVHN & MNIST-M & MNIST & Avg. 			\tabularnewline \hline 	
           
% 	Baseline & - & - &  - &78.6  &61.7 &58.8 &95.8 &73.7			\tabularnewline 
    				
	\hline 	 StandardAug   & Non-S & $\mathcal{A}$ &94.55&76.18&63.88&95.48&82.52
    \tabularnewline 
        
% 	\hline 	 TrivialAug   & A & A &93.95 &76.52 &65.32 &97.30 &83.27
%         \tabularnewline 
    				
    TrivialAug    & S  & $\mathcal{A}$  & 94.88 &80.65 &71.43 &98.35 &86.33			\tabularnewline 
	\hline 
	TripleE-Style (ours) & S   & $\mathcal{A}'$  &95.33  &81.75 &72.52 &98.28 &86.97		\tabularnewline 
					
	TripleE-Fourier (ours) & S  & $\mathcal{A}'$ &95.48  &81.22 &73.15 &98.33 &87.04
			\tabularnewline	
			
			\bottomrule
	\end{tabular}}	\label{tab:ablation_eaug}}
\end{table}

\begin{table}[!t]
	\centering
	\caption{Ablation studies on $m$ in EReplayD on Digits-DG.}
	\resizebox{\linewidth}{!}{{\begin{tabular}{c|cccc|c}	\toprule	
	
	Sub-sample times & SYN & SVHN & MNIST-M & MNIST & Avg. 			\tabularnewline \hline 	
    				
    $m=1$ \scriptsize{(\emph{without EReplayD})} &94.62 &79.83 &69.85 &97.43 &85.43 		\tabularnewline
					
	$m=2$ & 95.48  &79.75 & 68.7 & 98.08  & 85.50			\tabularnewline
					
	 $m=3$ & 95.33  & 81.75 & 72.52 & 98.28 & \textbf{86.97} \tabularnewline 
        				
       $m=4$   &94.87  &79.48  &70.67  &98.00  & 85.76
		\tabularnewline 

  $m=5$   &94.47  &78.10 &70.07  &97.98  &85.16  \tabularnewline

				\bottomrule
	\end{tabular}}
	\label{tab:ablation_m}}
\end{table}

\noindent \textbf{Ablation of EReplayB: a small batch with sufficient replay can enhance the generalization.}
Fig.~\ref{fig:ereplay} shows ablation study on parameter $b$ and $r$. Here, we empirically let $b$ equal to $r$ and show their effects to the performance. 
From Fig.~\ref{fig:ereplay} we can see that when $b = r = 4$ in Digits-DG and PACS, our method can achieve the best performance. The result verified the effectiveness of training a small batch (subset) with replay (augment for multiple times).
Whereas when $b = r = 32$, our method can achieve the best result on Office-Home. As $b$ and $r$ increase, our performance may be higher, but this experiment has a high computational cost and is left as our future work. 
% For the effects when $b \neq r$, please see the results in Appendix. 
%because we find that this setting has a better performance. We also do more experiments in the Appendix to show the situation that $b$ is not equal to $r$. 

\noindent \textbf{Ablation of ESAug: exhaustive and singular augmentation can boost the generalization.}
Table~\ref{tab:ablation_eaug} shows the effectiveness of our ESAug in terms of augmentation type and list. 
Compared with StandardAug, TrivialAug uses singular augmentation, which improves the performance from 82.52\% to 86.33\%. Therefore, we verified that a singular augmentation type effectively improves the performance. 
By comparing StandardAug and our method, we demonstrated that including a context-based image augmentation, Fourier or style, can further enhance the performance to 86.9\% and 87.0\%, respectively.

\noindent \textbf{Ablation of $m$ in EReplayD: episodically replay sub-datasets can help the generalization.}
Table~\ref{tab:ablation_m} shows the results with different sub-sample times in EReplayD. We tried $m$ from 1 to 5 and we found that setting $m$ to 3 can achieve the best performance. 
In particular, our method can achieve 1.5\% improvement over the model without EReplayD ($m=1$).

\noindent \textbf{Compare EReplayD with traditional ensemble.}
Table~\ref{tab:tradensemble} shows the effectiveness of EReplayD. A traditional ensemble refers to train $m$ networks on a whole dataset and ensemble the predictions from $m$ networks as the final result. From Table~\ref{tab:tradensemble} we can see that our proposed EReplayD, \ie, ensemble by periodically subsampling, outperforms the traditional model ensemble method by 1.33\%. The result verified the effectiveness of our main assumption, that is, to focus more on learning subsets (sub-datasets).

\begin{table}[!t]
	\centering
	\caption{Compare EReplayD with the traditional ensemble on Digits-DG. Note that each experiment ensembles three models.}
	\resizebox{\linewidth}{!}{{\begin{tabular}{c|cccc|c}	\toprule	
	
	Ensemble Type & SYN & SVHN & MNIST-M & MNIST & Avg. 			\tabularnewline \hline 	
    				
    Traditional Ensemble &94.93   &77.88  &71.83  &97.95  &85.64 \tabularnewline 
    				
    EReplayD (ours) & 95.33  & 81.75 & 72.52 & 98.28 & \textbf{86.97} \tabularnewline 

				\bottomrule
	\end{tabular}}	\label{tab:tradensemble}}
\end{table}

% \subsection{Visualization results}
% We use t-SNE~\cite{maaten2008visualizing} to visualize the learned embeddings for data from the target domain. 
% %All models are trained on \emph{SYN}, \emph{MNIST-M} and \emph{MNIST} and evaluated on \emph{SVHN} with the 4 conv layer backbone~\cite{zhou2020learning}.  
% As shown in Fig.~\ref{fig:tsne_digits}, compared to our baseline, our TripleE method better pulls together the embeddings \emph{within class} and enlarges the distances \emph{between classes}.
% The visualization demonstrates that our method learns much more structured embeddings and can generalize better.

% \begin{figure}[h]
% 	\centering
% 	\includegraphics[width=0.48\textwidth]{figure/tsne.jpeg}
% 	\caption{The t-SNE visualization on SVHN on Digits-DG. Each color represents a class. Bested viewed in color.} 
% 	\label{fig:tsne_digits}
% \end{figure}

\subsection{Visualization results}
\begin{figure}[htbp]
	\centering
	\includegraphics[width=0.48\textwidth]{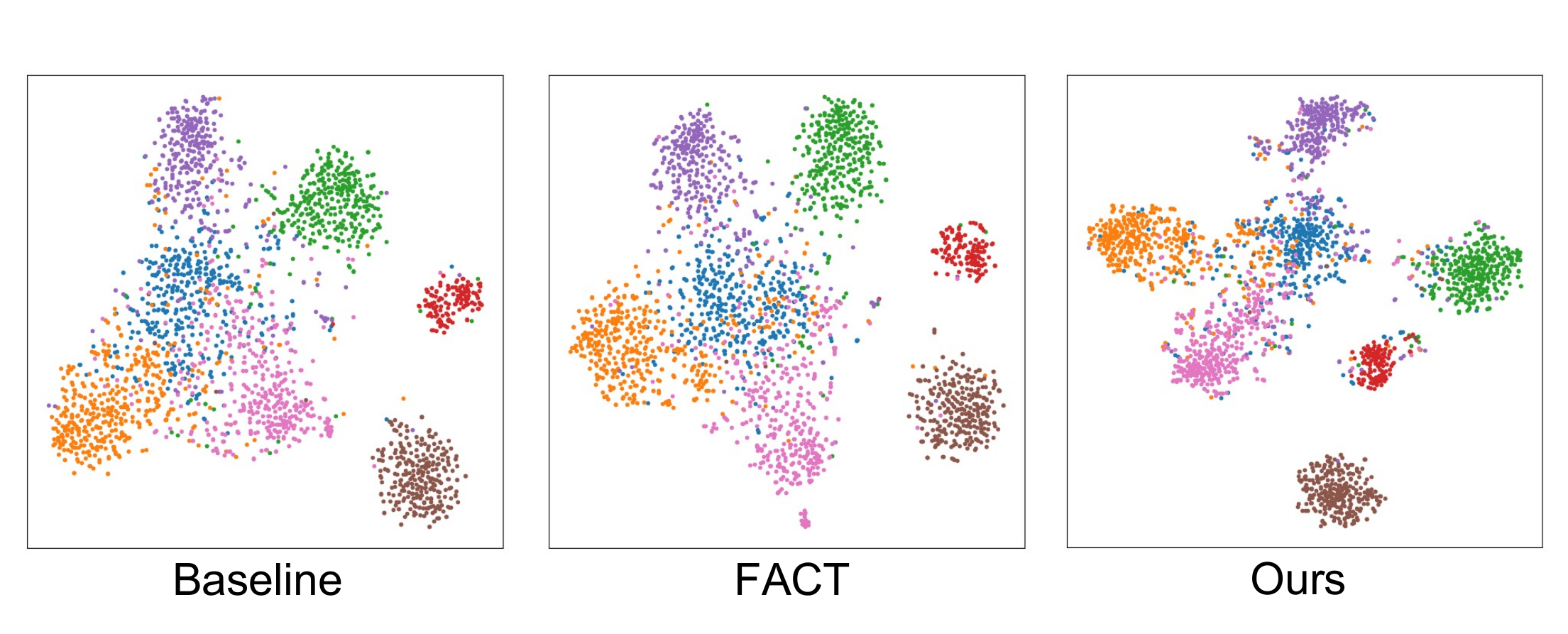}
	\caption{The t-SNE visualization on \emph{Cartoon} on PACS dataset. Each color represents a class. Bested viewed in color.} 
	\label{fig:tsne}
\end{figure}

\noindent \textbf{t-SNE visualization.} Fig.~\ref{fig:tsne} shows the visualization results of our baseline, FACT~\cite{xu2021fourier} and ours. The models are trained on \emph{art\_painting}, \emph{photo}, \emph{sketch} and tested on \emph{cartoon}. We can see that our method can better separate the classes on the unseen target domain than the other two methods. 

\begin{figure}[htbp]
	\centering
	\includegraphics[width=0.48\textwidth]{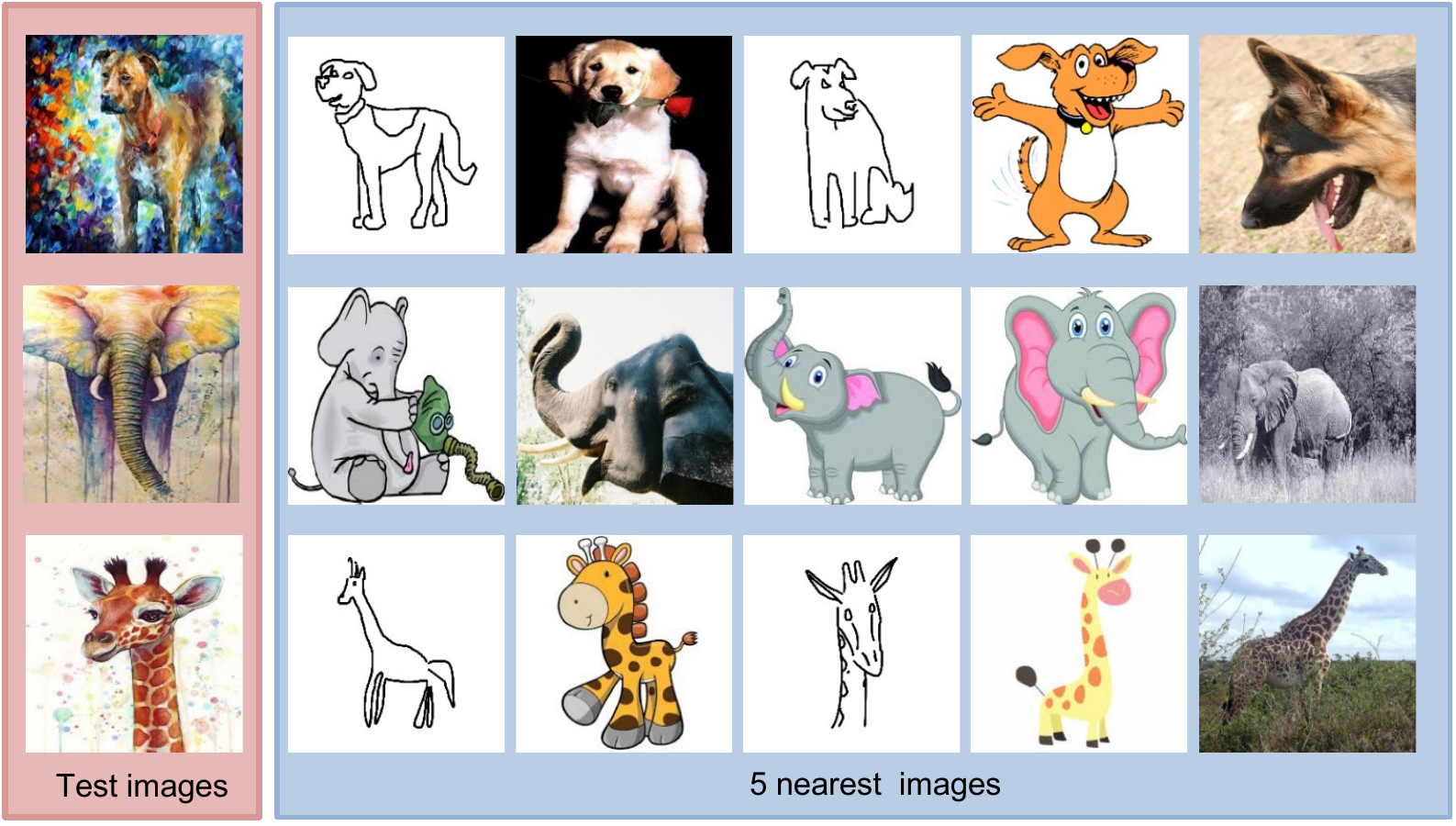}
	\caption{5-nearest neighbors for the images from unseen domain (\emph{art\_painting}). For each test image, the leftmost image is the closest image from the training dataset (\emph{sketch, photo and cartoon}). } 
	\label{fig:images}
\end{figure}

\noindent \textbf{$k$-nearest neighbors visualization.} Fig.~\ref{fig:images} shows the $k$-nearest neighbors ($k=5$) for the test image in the unseen domain. We can see that for a test image, the retrieved 5 nearest images from the training dataset have the same semantic information with the test image. The result demonstrates that our method can learn the domain-invariant features and has good model generalization ability.

\section{Discussion}

\subsection{Analysis on different $b$ and $r$ in EReplayB.}

Table~\ref{tab:ablation_br} shows the results with different $b$ and $r$ on Digits-DG. We can see that with a fixed batch size $b$, the performance improvement is limited when repeat times $r$ increase. The result shows that a high repeat time may not be necessary, and the result reaches the best performance when $r= 4$. 
On the other hand, when $r=4$, as $b$ increases, the performance worsens, and the best result is achieved at $b=4$ and $r=4$.

\begin{table}[t]
	\centering
	\caption{Effects of different $b$ and $r$ in EReplayB on Digits-DG.}
	\resizebox{\linewidth}{!}{{\begin{tabular}{c|cccc|c}	
	\toprule
	$b$ and $r$ &   SYN & SVHN & MNIST-M & MNIST & Avg.	\tabularnewline 
    \hline
   	
   	   $b=4,r=32$ &96.2  &79.57  &70.12  &98.30 &86.05
	\tabularnewline 
	   $b=4,r=16$ &96.45  & 80.02 & 70.38 & 98.42& 86.32
    \tabularnewline 
       $b=4,r=8$ &95.83  &78.60 &71.47 &98.37 & 86.07		
    \tabularnewline
   	$b=4,r=4$   & 95.48 & 81.22  &73.15 & 98.33 & 87.04
   	\tabularnewline 
   			
	\hline 

	$b=4,r=4$   & 95.48 & 81.22  &73.15 & 98.33 & 87.04
   			\tabularnewline 
   			
   $b=8,r=4$ &96.03  &78.45  &71.00  &98.58 &86.02
	\tabularnewline 
   $b=16,r=4$ &95.32  &79.57  &71.02  &98.22 &86.03
	\tabularnewline 
   $b=32,r=4$ &95.52  &79.20  &69.05  &98.30 &85.52
	\tabularnewline 

	\hline
	$b=4,r=4$ (\textbf{ours})  & 95.48 & 81.22  &73.15 & 98.33 & 87.04
	\tabularnewline	
	\bottomrule
	\end{tabular}}	
	\label{tab:ablation_br}}
\end{table}

\subsection{Effects of combining Fourier and stylization in ESAug.}
Table~\ref{tab:ablationAug} shows the results of combining Fourier and stylization in ESAug. We can see that adding both Fourier and Stylized transform achieves 86.76\% performance, indicating that these two context-based augmentations have similar effects for model generalization. Therefore, in this paper, we present the results generated by Fourier-based and stylized augmentation individually. 

\begin{table}[!t]
	\centering
	\caption{Effects of Fourier and Stylization in ESAug on Digits-DG.}
	\resizebox{\linewidth}{!}{{\begin{tabular}{c|cccc|c}	
	\toprule
	& SYN & SVHN & MNIST-M & MNIST & Avg.	\tabularnewline 
	\hline 	
    				
   Combine Fourier and Stylized  &95.45  &79.68  &73.5  &98.43  &86.76
    \tabularnewline

	TripleE-Fourier   & 95.33 & 81.75  &  72.52 & 98.28 & 86.97
	\tabularnewline 

	TripleE-Stylized   & 95.48 & 81.22  &73.15 & 98.33 & 87.04
	\tabularnewline	
	\bottomrule
	\end{tabular}}	
	\label{tab:ablationAug}}
\end{table}

\subsection{Effects of different augmentation probabilities in ESAug}
We include Fourier-based augmentation/ stylized augmentation into the standard augmentation list, consisting of 14 transformation functions. We studied different probabilities to include these context-exchange augmentations. 
Table~\ref{tab:ablationPro} shows that concatenating the context-exchange augmentation (Fourier/Stylized) with the standard augmentation list achieves the best overall performance. 

\begin{table}[!t]
	\centering
	\caption{Effects of different augmentation probabilities in EAug on Digits-DG.}
	\resizebox{\linewidth}{!}{{\begin{tabular}{c|c|cccc|c}	
	\toprule
	& Aug prob. & SYN & SVHN & MNIST-M & MNIST & Avg.	\tabularnewline 
	\hline

    Fourier &1/2  &95.47  &76.97  &69.37  & 98.03 &84.96
    \tabularnewline 
    stylized   & 1/2&95.75  &75.67  &69.53  &98.08  &84.76
    \tabularnewline

	\hline
	Fourier(ours)   & 1/15 & 95.33 & 81.75  &  72.52 & 98.28 & 86.97
	\tabularnewline 
	stylized(ours)   & 1/15 & 95.48 & 81.22  &73.15 & 98.33 & 87.04
	\tabularnewline 
	\bottomrule
	\end{tabular}}	
	\label{tab:ablationPro}}
\end{table}

%% file: sections/6.Conclusion.tex
% The key assumption is that an ideal DG model should learn consensus features
% and should not be affected by biased samples in source domains.
% Hence, we aim to derive a consensus model from
% source domains that can achieve good performance on target
% domains. Our idea is to regard data in source domains as
% biased samples and to derive consensus models by random
% sampling. To this end, our insight is to encourage the network
% to \textbf{\textit{focus on training on subsets (learning with replay)}}
% and \textbf{\textit{enlarge the data space in performing replays}}.

\noindent \textbf{Data availability} The six datasets analyzed during the current study are available on the following websites: PACS~\footnote{\url{ https://drive.google.com/drive/folders/1SKvzI8bCqW9bcoNLNCrTGbg7gBSw97qO}}, Digits-DG~\footnote{\url{https://drive.google.com/uc?id=15V7EsHfCcfbKgsDmzQKj_DfXt_XYp_P7}}, Office-Home~\footnote{\url{https://drive.google.com/file/d/0B81rNlvomiwed0V1YUxQdC1uOTg/view?resourcekey=0-2SNWq0CDAuWOBRRBL7ZZsw}}, VLCS~\footnote{\url{https://drive.google.com/uc?id=1skwblH1_okBwxWxmRsp9_qi15hyPpxg8}}, TerraInc~\footnote{\url{https://lilablobssc.blob.core.windows.net/caltechcameratraps/eccv_18_all_images_sm.tar.gz}}, DomainNet~\footnote{\url{http://ai.bu.edu/M3SDA/}}.

\section{Conclusion}
This paper presents a frustratingly easy yet should-know method for domain generalization. Unlike prior work that relies on a particular module, loss function, or structure design, our method is based on simple augmentations and ensembling. 
We reveal the key secret for model improvement is to encourage the network to train on subsets and enlarge the data space in learning on subsets. 
Our TripleE achieves this goal through three components: EReplayB, EReplayD, and ESAug. 
Experiments demonstrated that TripleE is outperforming or comparable to the state of the art, showing it could serve as a stepping stone for future research in domain generalization.

% \textbf{Acknowledgements} To be decided later. 